\documentclass[acmsmall, authorversion, nonacm, screen]{acmart}
\AtBeginDocument{%
  }

\acmJournal{TOMM}

\usepackage{color,soul}
\usepackage{subcaption}
\usepackage{threeparttable}

\begin{document}


\title[Summarization of Multimodal Presentations with Vision-Language Models]{Summarization of Multimodal Presentations with Vision-Language Models: Study of the Effect of Modalities and Structure}

\author{Théo Gigant}
\orcid{0009-0003-6392-8519}
\affiliation{%
  \institution{Université Paris-Saclay, CNRS, CentraleSupelec, Laboratoire des signaux et systemes}
  \city{Gif-Sur-Yvette}
  \country{France}
}
\email{theo.gigant@l2s.centralesupelec.fr}

\author{Camille Guinaudeau}
\affiliation{%
  \institution{Université Paris-Saclay, CNRS, LISN}
  \city{Orsay}
  \country{France}
}
\orcid{0000-0001-7249-8715}

\author{Frédéric Dufaux}
\affiliation{%
  \institution{Université Paris-Saclay, CNRS, CentraleSupelec, Laboratoire des signaux et systemes}
  \city{Gif-Sur-Yvette}
  \country{France}
}
\orcid{0000-0001-6388-4112}


\begin{abstract}

Vision-Language Models (VLMs) can process visual and textual information in multiple formats: texts, images, interleaved texts and images, or even hour-long videos.
In this work, we conduct fine-grained quantitative and qualitative analyses of automatic summarization of multimodal presentations using VLMs with various representations as input. From these experiments, we suggest cost-effective strategies for generating summaries from text-heavy multimodal documents under different input-length budgets using VLMs.
We show that slides extracted from the video stream can be beneficially used as input against the raw video, and that a structured representation from interleaved slides and transcript provides the best performance.
Finally, we reflect and comment on the nature of cross-modal interactions in multimodal presentations and share suggestions to improve the capabilities of VLMs to understand documents of this nature.

\end{abstract}






\maketitle

\section{Introduction}

Presentations play a central role in academic, business, and technical communication, serving as a widespread means of conveying information. By leveraging multimodal elements such as visual slides and speech, presentations offer a dynamic and engaging way to present ideas and data.
Their sheer volume and the frequency with which individuals encounter them make summarization both a practical and a pressing need. Summaries of multimodal presentations allow users to efficiently extract key points, navigate the presentation, and improve retention, without requiring in-depth engagement with the full presentation.

\begin{figure}[h]
    \centering
    \includegraphics[width=\linewidth]{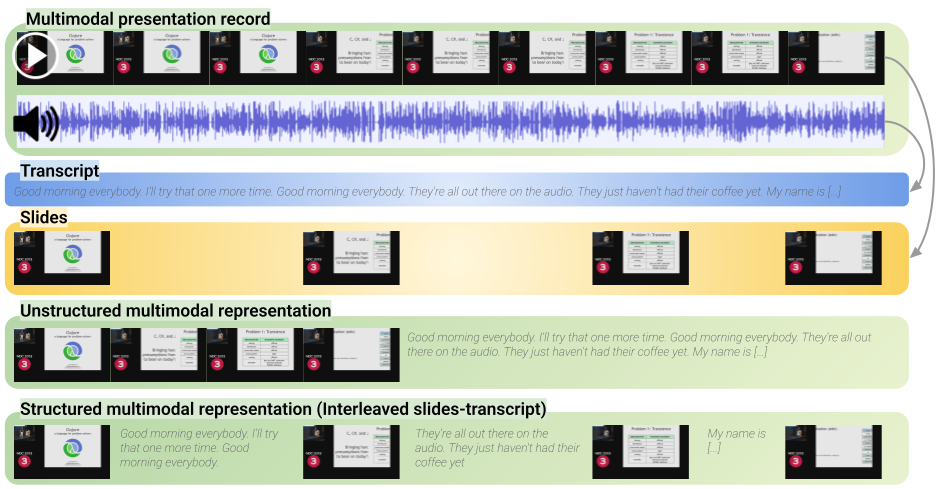}
    \caption{VLMs are able to process a multimodal presentation in various unimodal and multimodal representations.}
    \label{fig:input-settings}
\end{figure}

Presentations are multimodal when they include a visual medium, typically a slideshow, in addition to the speech. Meetings, lectures, and conferences commonly follow this format.
Multimodal presentation records usually consist of text-heavy video and audio streams. A transcript and key images that span the entire video can be extracted from these records \cite{gigant_tib_2023}. Using these extracted modalities or the original modalities, one can construct representations of the presentation to be used by VLMs to perform tasks such as summarization. Examples of such unimodal and multimodal representations for a presentation are illustrated in Figure \ref{fig:input-settings}. The choice of representation will have different token lengths and result in different downstream performances and costs.


VLMs are usually trained and evaluated on short input, including text and a few images, or short videos \cite{alayrac_flamingo_2022, liu_visual_2023}. Previous works have highlighted the poor performance of these models with a longer input \cite{wu_longvideobench_2024}. Although long context performance on synthetic benchmarks can be inherited from the underlying language model \cite{zhang_long_2024}, they usually do not reflect performance in real-world scenarios \cite{li_long-context_2024}.
The raw video representation can be too long and lie significantly outside of the training context size for these models, rendering summarization very challenging \cite{wang_qwen2-vl_2024}. However, we can build a shorter interleaved image-text representation resembling the training data of some VLMs \cite{alayrac_flamingo_2022,  laurencon_what_2025, laurencon_building_2024}.

In this paper, we examine how the choice of representation of multimodal presentations affects their associated cost and performance for VLM-based summarization.
We specify different representations for the model, using single or multiple modalities, including slides, text extracted from slides, the entire video, and the transcribed audio, optionally structured by using the time-alignment between these modalities. Furthermore, we perform a fine-grained analysis of the performance of Qwen2-VL \cite{wang_qwen2-vl_2024} with these different input representations to understand the impact of the modalities and the structure information on the generated summaries. Additionally we benchmark multiple recent state-of-the-art open-weights VLMs using an interleaved slides-transcript representation. 

The contributions of this paper can be summarized as follows:
\begin{itemize}
    \item We benchmark open-weights VLMs for summarization of multimodal presentations using an interleaved slides-transcript input.
    \item We conduct a fine-grained quantitative and qualitative analysis of the cost and performance associated with many unimodal and multimodal input representations in Qwen2-VL.
    \item We suggest cost-effective representation strategies for summarization of multimodal presentations with VLMs.
    \item We discuss the challenges emerging from using multimodal representations in VLMs and share suggestions to mitigate the current shortcomings.
\end{itemize}

\section{Related Work}

\subsection{Abstractive Summarization}

Previous work has explored various approaches to abstractive summarization, \textit{i.e.} efficiently condensing documents in short generated textual summaries. Recently, models based on the transformer encoder-decoder architecture \cite{vaswani_attention_2017} such as Bart \cite{lewis_bart_2020} or T5 \cite{raffel_exploring_2020} have shown great performance for abstractive summarization of short textual documents.

These architectures have been adapted for long input, by interchanging the attention module with linear scaling alternatives such as local and global attentions \cite{ainslie_etc_2020, guo_longt5_2022, beltagy_longformer_2020, zaheer_big_2020}.

Current Large Language Models (LLMs), based on the transformer decoder architecture, perform multiple tasks, including abstractive summarization, with remarkable success \cite{zhang_systematic_2024}. Their capabilities even scale to a very large input due to efficient attention implementations \cite{dao_flashattention_2022, dao_flashattention-2_2023} and context extension methods \cite{peng_yarn_2023}.

However, these methods fall short for summarization of multimodal presentations because they neglect the visual modality, which is often critical to convey information in such contexts.

\subsection{Vision-Language Models and Video-Language Models}

Vision-Language Models (VLMs) \cite{alayrac_flamingo_2022} are early-fusion multimodal models comprised of a vision encoder \cite{radford_learning_2021, zhai_sigmoid_2023} connected to a pretrained LLM. They carry out the document understanding and instruction-following capabilities of LLMs to visual or multimodal documents \cite{liu_visual_2023}.

The training data for recent VLMs include a significant amount of text-heavy images \cite{laurencon_what_2025} such as document OCR or document visual question answering. VLMs can also be trained with interleaved text and images \cite{alayrac_flamingo_2022}, meaning that they can process an arbitrary number of images in between text tokens.

Some VLMs can also process video input on top of images, optionally using a dedicated video encoder \cite{lin_video-llava_2024}.
Video VLMs are usually trained with minutes-long \textit{in-the-wild} videos picturing little to no text \cite{luo_valley_2025}. In contrast, the video stream of multimodal presentation records is typically long, redundant, and text-heavy. As such, summarization of multimodal presentations lies out of the training domain of video VLMs, and, by extracting key frames, can be made more similar to the domain for recent VLMs using the interleaved images-text format \cite{alayrac_flamingo_2022, laurencon_what_2025, laurencon_building_2024}.

\citet{wu_longvideobench_2024} highlighted the low performance of open-weights VLMs for long-context interleaved video-language understanding.
\citet{zhang_long_2024} suggested that VLMs can inherit long-context performance from their underlying LLMs, exemplifying on a synthetic benchmark. However, \citet{li_long-context_2024} showed that the performance of LLMs on synthetic benchmarks does not reflect real-world performance.


\subsection{Abstractive Summarization of text-heavy long multimodal documents}

The summarization of long videos \cite{alaa_video_2024, apostolidis_video_2021}, and the understanding of documents with a text-heavy visual modality \cite{kim_ocr-free_2022} using VLMs have received significant attention in recent research. However, the intersection of these two domains remains relatively understudied.
Some early efforts \cite{gigant_tib_2023, liu_what_2025} have benchmarked approaches using single-modality or out-of-domain baselines, revealing critical limitations of existing methods in effectively performing abstractive summarization of text-heavy long multimodal documents, such as those found in multimodal presentations.

TIB \cite{gigant_tib_2023} is a dataset containing 9,103 multimodal scientific presentations of multiple domains, with abstracts as reference summaries. Each presentation was processed to extract the transcript and key frames. The transcripts were used to benchmark text-only abstractive summarization models, and key frames were provided to use with upcoming long-context VLMs.

VISTA \cite{liu_what_2025} is a similar benchmark, consisting of 18,599 records of multimodal presentations from AI conferences, with their abstracts. The authors tested LLMs with transcript or OCR input, as well as VLMs with video input, and advocated for explicit planning to improve the quality and factual consistency of summaries.

Compared with these, our work constitutes a more comprehensive investigation into VLMs' ability to generate summaries from multimodal content, with a focus on the effect of input document representation.

\section{Methodology}

\subsection{Dataset}
We constitute a benchmark derived from the TIB dataset \cite{gigant_tib_2023}, initially comprised of 9103 records of scientific presentations, with transcribed speech, extracted key frames and author-provided abstract. We chose TIB over VISTA because of the availability of pre-processed key frames and transcripts, and the longer length of the presentations.

The original TIB dataset contains reference summaries that were author-provided, so the abstract style, quality, and content are varied. In order to create a higher-signal subset, we filter the top quality abstracts using a LLM-as-a-judge method.
Based on the title and abstract, we prompt SmolLM2 \cite{allal_smollm2_2025} to score the abstracts from 1 to 9 and keep only the references scoring 9.

\citet{gigant_tib_2023} stated that the extracted key frames in the TIB dataset often contain screen-recorded or live-captured slideshows, but can also be a live recording of the speaker.
To filter out the latter, we use SigLip \cite{zhai_sigmoid_2023} to perform zero-shot classification on key frames to identify those that contain a slide and those that do not. We discard the key frames that are not slides, and the presentations with less than 3 slides after this stage.


After processing, our benchmark consists of 822 presentations, including transcribed speech and slides with time codes. Using these time codes, we can create an interleaved sequence of slides followed by transcription of the simultaneous speech. For the rest of this work, we will refer to this format as "interleaved" or "structured". Otherwise, we default to a sequence that includes all slides, followed by the transcribed speech. We call this default representation "unstructured".

We share the filtered benchmark\footnote{\url{https://huggingface.co/datasets/gigant/tib-bench}}, as well as a tool to visualize the interleaved representation of the multimodal presentations\footnote{\url{https://huggingface.co/spaces/gigant/slide-presentation-viz}} on the Hugging Face Hub.

\subsection{Models}

For a fine-grained analysis of the task at hand, we consider open-weights VLMs that fit on a single A-100 GPU, and are able to process multiple images, interleaved images-text format, videos and long-context input.
The outputs are sampled with greedy decoding, with a maximum output size of 1024 tokens.
For all models, we adapt the same prompt, using the following instructions: 

\texttt{Provide a summary of the presentation in a short paragraph, based on the transcript and the slides. The summary should introduce the topic, and the speaker and the presentation context if relevant. It will be written to be informative about the content for a prospective attendee. \{document\}}

The main model we are studying in this work is Qwen2-VL \cite{wang_qwen2-vl_2024}, as it checks all the requirements and also allows for arbitrary visual resolution. It also comes in different sizes; specifically, we work with the 2B and 7B instruction-tuned checkpoints.
All of these characteristics make it convenient to rely on Qwen2-VL to perform a fine-grained analysis of the impact of the input modalities and structure on the generated summaries.
In the benchmark, we include other VLMs capable of handling our multimodal presentations, namely Idefics-2 \cite{laurencon_what_2025}, Idefics-3 \cite{laurencon_building_2024}, and Phi-3.5-vision \cite{abdin_phi-3_2024}.


We believe that the findings of this work can be generalized to other models, due to the similarity in architecture and training data between current VLMs \cite{mckinzie_mm1_2025, laurencon_what_2025}.

\subsection{Extractive Statistics}

We use the transcripts provided in the TIB dataset, and extract words from the slides using an Optical Character Recognition (OCR) model: GOT-OCR2.0 \cite{wei_general_2024}. This allows to compute word-based statistics and metrics with respect to the speech by using the transcript, and to the slides by using the OCR, for modality-specific measures of extractiveness and relevance.


\begin{figure}[h]
    \centering 
    \includegraphics[width=0.7\linewidth]{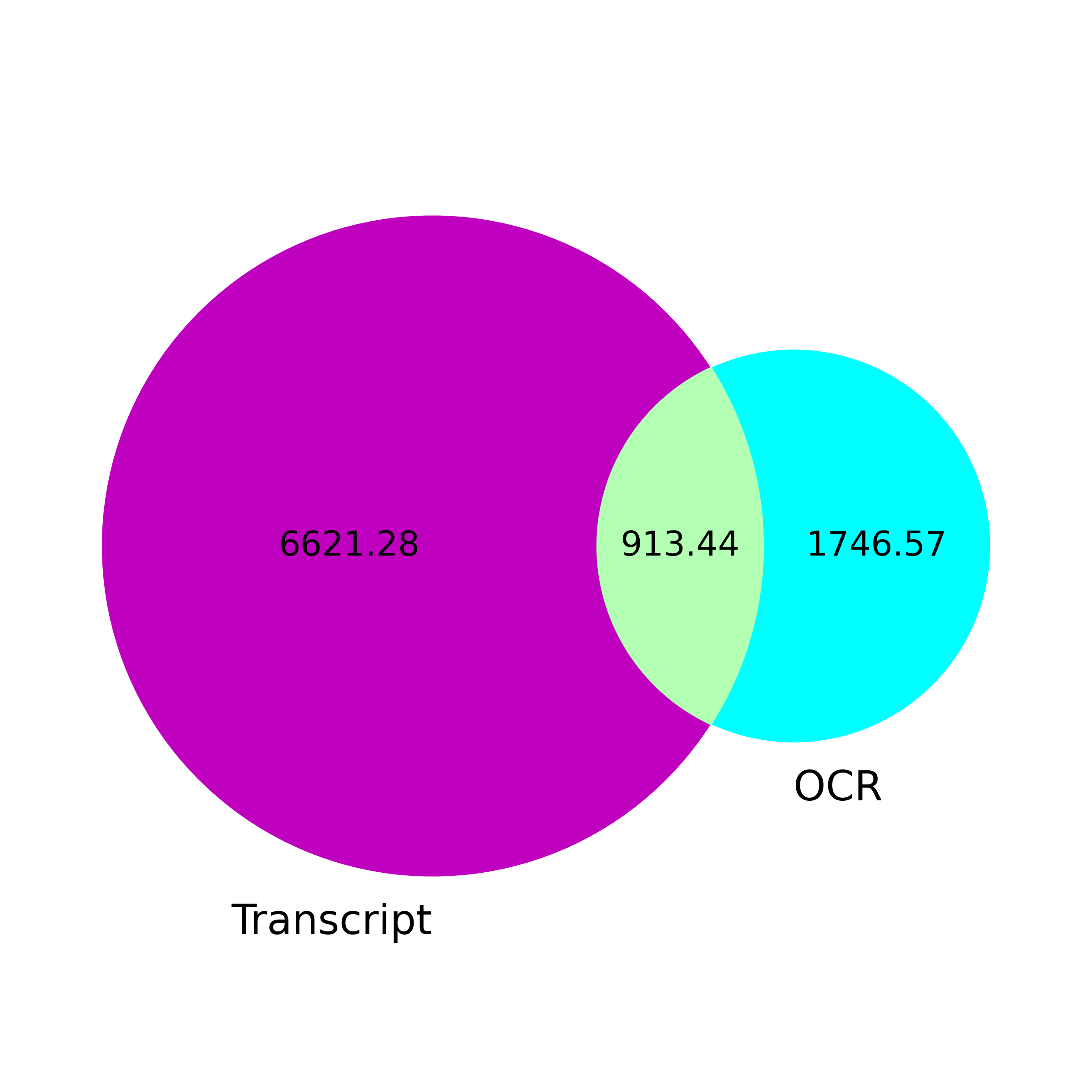}
    \caption{Average token count in speech transcript and OCR, and their overlap}
    \label{fig:venn-token-count}
\end{figure}

We compute the average token count in the transcripts, the OCR, and the vocabulary overlap between the two. These statistics are illustrated in Figure \ref{fig:venn-token-count}. We remind the reader that transcriptions from the images and the speech are imperfect and, as a consequence, the overlap might be inaccurate.

The extractive statistics of generated summaries are computed in multiple settings, with respect to the text extracted from the slides, the speech, or their concatenation, called \texttt{ocr}, \texttt{transcript} and \texttt{overall} respectively.

The extractive coverage refers to the vocabulary overlap between the source and the summary. The extractive density is the average length of the extractive fragments from the source in the summaries.
A more extractive summary with respect to a source, can have a higher vocabulary overlap \textit{ie} higher coverage, and/or longer extractive fragments, \textit{ie} higher density.


\subsection{Evaluation Metrics}

\label{sec:eval-metrics}

Many metrics have been proposed for the fine-grained evaluation of summarization systems. However, meta-evaluations have shown that they usually scale poorly for long document summarization \cite{koh_how_2022}. Metrics such as Bart Score \cite{yuan_bartscore_2021} and ESTIME \cite{vasilyev_estime_2021} rely on encoding the input document in a pretrained model, which has a limited input size, usually less than $1000$ tokens. The documents in our experiments exceed on average more than ten times this input size, so we believe that these scores will not be reliable in this setting.

The Rouge score \cite{lin_rouge_2004} is the standard metric for the evaluation of summaries. It gives an accurate estimate of the overall performance of a system. Rouge-$n$ counts the $n$-gram overlaps between the proposed summary and a reference summary. We will work primarily with unigram and bigram-based Rouge scores, denoted Rouge-1 ($R_1$) and Rouge-2 ($R_2$).

GRUEN ($G$) \cite{zhu_gruen_2020} is a reference-free metric intended to evaluate the linguistic quality of generated texts. It only uses the generated text as input, so it does not need to account for the input being long and multimodal.

The importance-based relevance score ($IbR$) \cite{gigant_mitigating_2024} has been designed to provide a reference-free evaluation of the relevance of the summaries. The relevance of a summary is
the measure of whether it contains the
main ideas from the source document. As this metric relies on word importance scoring based on information retrieval methods, it scales well with long input and has been shown to correlate well with human-evaluated relevance of summaries on long-document summarization datasets.
It uses the source as a textual reference and will have to be adapted for use with multimodal input. To this end, we are using extracted text from the speech and/or slides as sources. Similarly to the extractive statistics, we define three modes for this metric, in accordance to the source input, called \texttt{ocr}, \texttt{transcript} and \texttt{overall}.
These input-modality-specific metrics allow us to conduct a more fine-grained analysis of the generated summaries with respect to the input modalities.

The Rouge scores ($R_1$ and $R_2$) provide us with an estimate of the overall performance of the summarization systems. GRUEN ($G$) scores the linguistic qualities of the systems. Finally, the multimodality-adapted importance-based relevance scores ($IbR_{\textit{\tiny{transcript}}}$, $IbR_{\textit{\tiny{ocr}}}$ and $IbR_{\textit{\tiny{overall}}}$) give us an estimate of how well the model gathers the important information from the document, attributed to each input modality.


\section{Benchmark}

We benchmark some VLMs suited for the multimodal summarization task.
The selected models are instantiated in the \texttt{bfloat16} format if possible; otherwise, we use 4-bit quantization and optionally truncate the input to 16k tokens to avoid out-of-memory errors.

The VLMs we are benchmarking are the following:

\begin{itemize}
    \item Idefics-2 \cite{laurencon_what_2025} is based on the Mistral-7B LLM, connected to the Siglip image encoder. It was trained with context length up to 2k tokens.
    \item Idefics-3 \cite{laurencon_building_2024} is based on the Llama-3 LLM, connected to the Siglip image encoder. It was trained with a context length up to 10K tokens.
    \item Phi-3.5-vision \cite{abdin_phi-3_2024} is based on the Phi-3.5-mini LLM, connected to the Clip image encoder. It was trained with a context length up to 128k tokens.
    \item Qwen2-VL \cite{wang_qwen2-vl_2024} is based on the Qwen2 LLM, connected to a custom image encoder. It was trained with a context length up to 16k tokens.
\end{itemize}

We evaluate the summaries generated by these models with the collection of metrics described in Section \ref{sec:eval-metrics}.
The results are reported in Table \ref{tab:benchmark-vlms}.

\begin{table}[h]
    \centering
    \caption{Results of open-weight VLMs on TIB-bench}
    \begin{tabular}{ccp{0.7cm}cccccc}
    \hline
        Model & \tiny{\#Params} & \tiny{Visual \par token \par budget} & $R_{1} \uparrow$ & $R_{2}\uparrow$ & $G\uparrow$ & $IbR_{\texttt{\tiny{transcript}}}\uparrow$ & $IbR_{\texttt{\tiny{ocr}}}\uparrow$ & $IbR_{\texttt{\tiny{overall}}}\uparrow$\\
        \hline
        Idefics-2 \cite{laurencon_what_2025} & 8B & 64 & 4.9 & 0.3 & 36.1 & 1.2 & 0.6 & 1.2 \\
        Idefics-3$^\texttt{q}$ \cite{laurencon_building_2024} & 8B & 169 & 12.4 & 1.9 & 48.0 & 21.2 & 7.9 & 16.3 \\
        Phi-3.5-vision$^{\texttt{q} \dagger}$ \cite{abdin_phi-3_2024} & 4B & 144 & 17.1 & 2.9 & 51.6 & 29.7 & 8.9 & 20.2 \\
        Qwen2-VL \cite{wang_qwen2-vl_2024} & 2B & 512 & \textbf{27.1} & \textbf{5.4} & \textbf{57.1} & \textbf{53.8} & \textbf{17.2}  & \textbf{33.4} \\
        \hline
        
    \end{tabular}
    \label{tab:benchmark-vlms}
    \begin{tablenotes}
    \centering
    \item \tiny{
    $^\texttt{q}$ denotes models with weights quantized at 4-bits}
    \item \tiny{
    $^\dagger$ denotes models whose input was truncated to 16k tokens}

    \end{tablenotes}
\end{table}

Idefics-2 has been trained with an interleaved image-text format that does not exceed 2k tokens. Even if it is technically capable of processing longer documents, we observe an inability to generate coherent summaries with significantly longer input. In fact, the GRUEN ($G$) score shows that the linguistic quality of the summaries generated by Idefics-2 is low, revealing the difficulty of this model in producing coherent text in such long context situations.

The other models are trained on longer context and all produce summaries of higher linguistic quality. The relevance of the generated summaries, as measured by the importance-based relevance ($IbR$) scores, highlights the differences between these models in the understanding of the presentations.

From Idefics-2 to Idefics-3 multiple changes can explain the relative improvement in this task, including a longer context training, a more capable backbone LLM, and more visual tokens per image.
\citet{laurencon_what_2025} observed no gain from adding visual tokens per image after 64, while \citet{laurencon_building_2024} observed that the OCR task benefited from more visual tokens per image.
Similarly to OCR, the summarization of multimodal presentations is based on text-heavy images. In Section \ref{sec:visual_scaling}, we investigate the scaling of performance with more visual tokens per image to see if these observations translate to this task.

Even with a smaller parameter count, Phi-3.5-vision scores significantly better than Idefics-3, most likely due to its training data including videos and more long context examples.
Between Idefics-3 and Phi-3.5-vision, the difference in performance seems to be linked to the language model more than the vision encoder, probably because of better long context understanding.

Qwen2-VL scores even better than Phi-3.5-vision, with a smaller parameter count, likely due to various improvements including a larger visual token budget and a larger and more diverse training dataset. Contrary to Phi-3.5-vision, it can process the whole non-truncated input with non-quantized weights.

The importance-based relevance ($IbR$) scores reveal that with these models, visual and textual understanding are improved jointly.
A new VLM release usually relies simultaneously on better training data and a better backbone LLM, causing both visual and language understanding to improve.

For all these models, the task of summarization of multimodal presentations is most likely out-of-domain, which would benefit models such as Qwen2-VL, trained on a broad range of tasks, including analogous ones such as long document and video understanding.

In the following experiments, we are further exploring the capabilities of Qwen2-VL and using its video understanding capabilities and dynamic visual token budget to provide a fine-grained analysis of the impact and scaling of different visual and multimodal settings.

\section{Fine-grained Analysis}

\subsection{Impact of the structure and modalities}

Qwen2-VL \cite{wang_qwen2-vl_2024} can process text, images, and/or videos as input. It includes a dynamic resolution mechanism that enables it to process images of varying resolution into different numbers of visual tokens. We experiment with different numbers of visual tokens, and call "visual token budget" the maximum number of visual tokens per image. The whole budget is used for square images of sufficient resolution; in other cases, the actual number of visual tokens used is smaller.

In the following experiments, visual token budget varies from 2 to 512. Given the various shapes and resolutions of the slides, the actual number of visual tokens per image is dynamic and ranges on average from 1 to 183. The measured averages are reported in Table \ref{tab:visual_tokens}. At 512 tokens per image, we already exceed the maximum resolution of the extracted slides in the dataset, explaining the larger relative gap between the budget and the actual visual tokens used. The videos are kept at the original resolution, which is higher on average.

\begin{table}[h]
    \centering
    \caption{Measured average number of visual tokens for different visual token budgets, and resulting input length for slides and video sampled at 0.1 or 1 fps. Settings where the input length exceeds the model context size are italicized.}
    \label{tab:visual_tokens}
    \begin{tabular}{p{0.15\linewidth}p{0.15\linewidth}p{0.15\linewidth}p{0.15\linewidth}p{0.15\linewidth}}
    \hline
        Visual token \par budget & Average visual \par tokens per \par slide & Total tokens \par (slides) & Total tokens \par (video@0.1fps) & Total tokens \par (video@1fps) \\
        \hline
        2 & 1.0 & 0.1k & 0.1k & 1k \\
        8 & 6.0 & 0.2k & 0.6k & 6k \\
        64 & 50.5 & 1k & 3k & \textit{31k} \\
        128 & 119.5 & 4k & 13k & \textit{134k} \\
        512 & 182.8 & 6k & \textit{53k} & \textit{538k} \\
        \hline
    \end{tabular}
\end{table}


In slide-based multimodal presentations, the video stream is highly redundant and consists mostly of slides spanning time segments to cover it. We conjecture that the visual stream can be preprocessed to keep only a few key frames without significant loss of information. The method for obtaining these key frames is described in the paper that introduces the TIB dataset \cite{gigant_tib_2023} and focuses on retrieving the slides from the video stream.

In order to validate this assumption, we feed the model with either the full video or the extracted key frames.
We conduct experiments with different sample rates: 0.1 frames per second, following \citet{liu_what_2025}, or 1 frame per second. The average input token length rapidly grows very large with more tokens per frame, as detailed in Table \ref{tab:visual_tokens}.

\begin{figure}[h]
    \centering
    \includegraphics[width=0.7\linewidth]{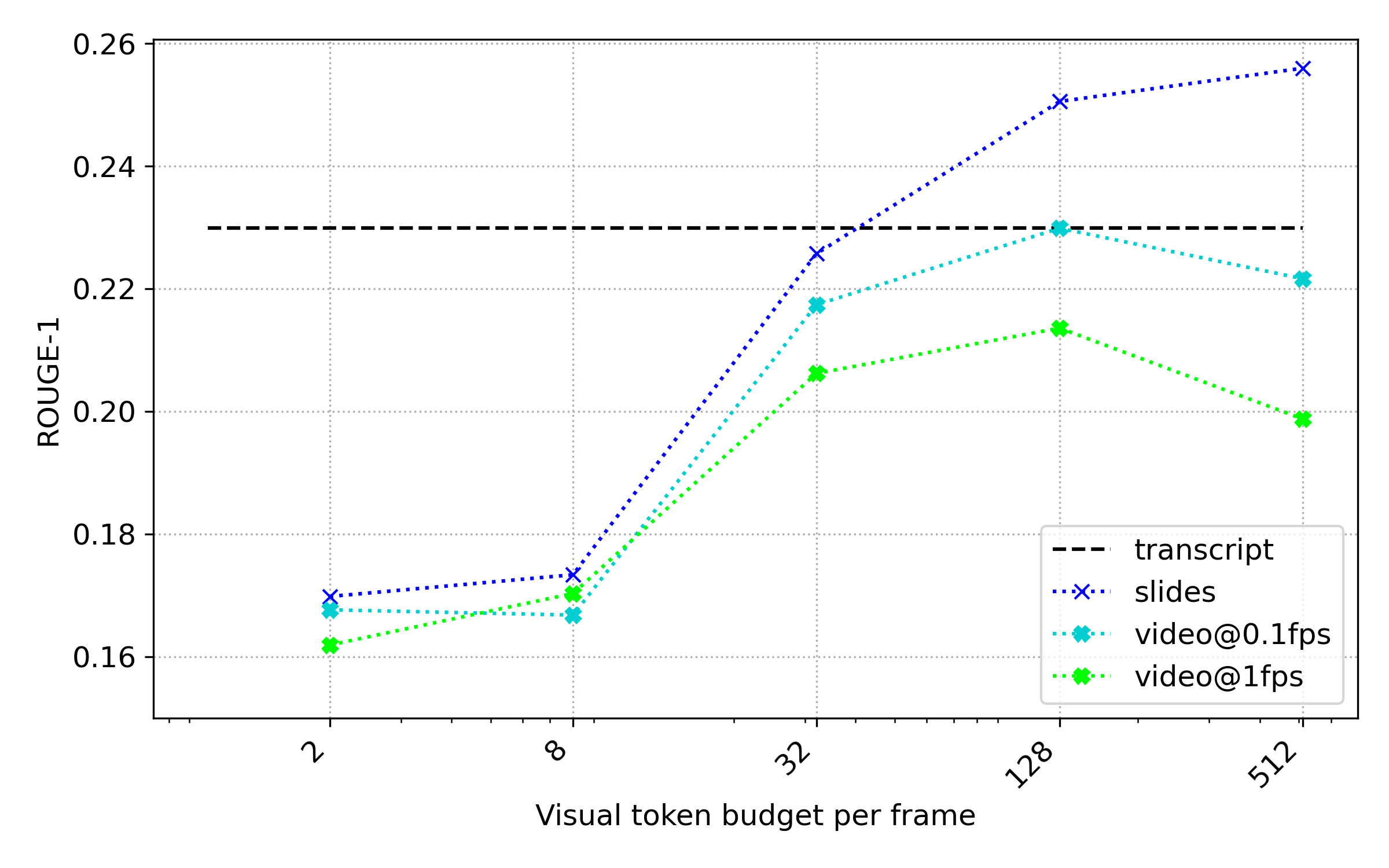}
    \caption{Unimodal performance at different token budgets}
    \label{fig:slide_v_video}
\end{figure}

Figure \ref{fig:slide_v_video} shows the performance of the model with input of a single modality: video, slides or transcript. At all visual token budgets per frame, the slides input outperforms the video input, all while consisting of much fewer frames (\textit{cf} Table \ref{tab:visual_tokens}).
The use of key frame input demonstrates a notable advantage with significantly improved performance and substantially lower computational cost, as evidenced in Table \ref{tab:visual_tokens}. 
The key frame input with high per slide budget also compares favorably with the transcript input.


\label{sec:structure}

As a way to represent the multimodal input for the model, we consider two options: concatenating the text after the slides or interleaving the tokens according to the temporal information gathered during preprocessing. In the interleaved format, the visual tokens for a slide are directly followed by the tokens of the transcribed simultaneous speech. Thus the images are near the related text in the sequence.
The other option is a sequence of all slides followed by the transcribed speech. In this default representation, we do not leverage any knowledge about the structure of the presentation.

\begin{figure}[h]
    \includegraphics[width=0.7\linewidth]{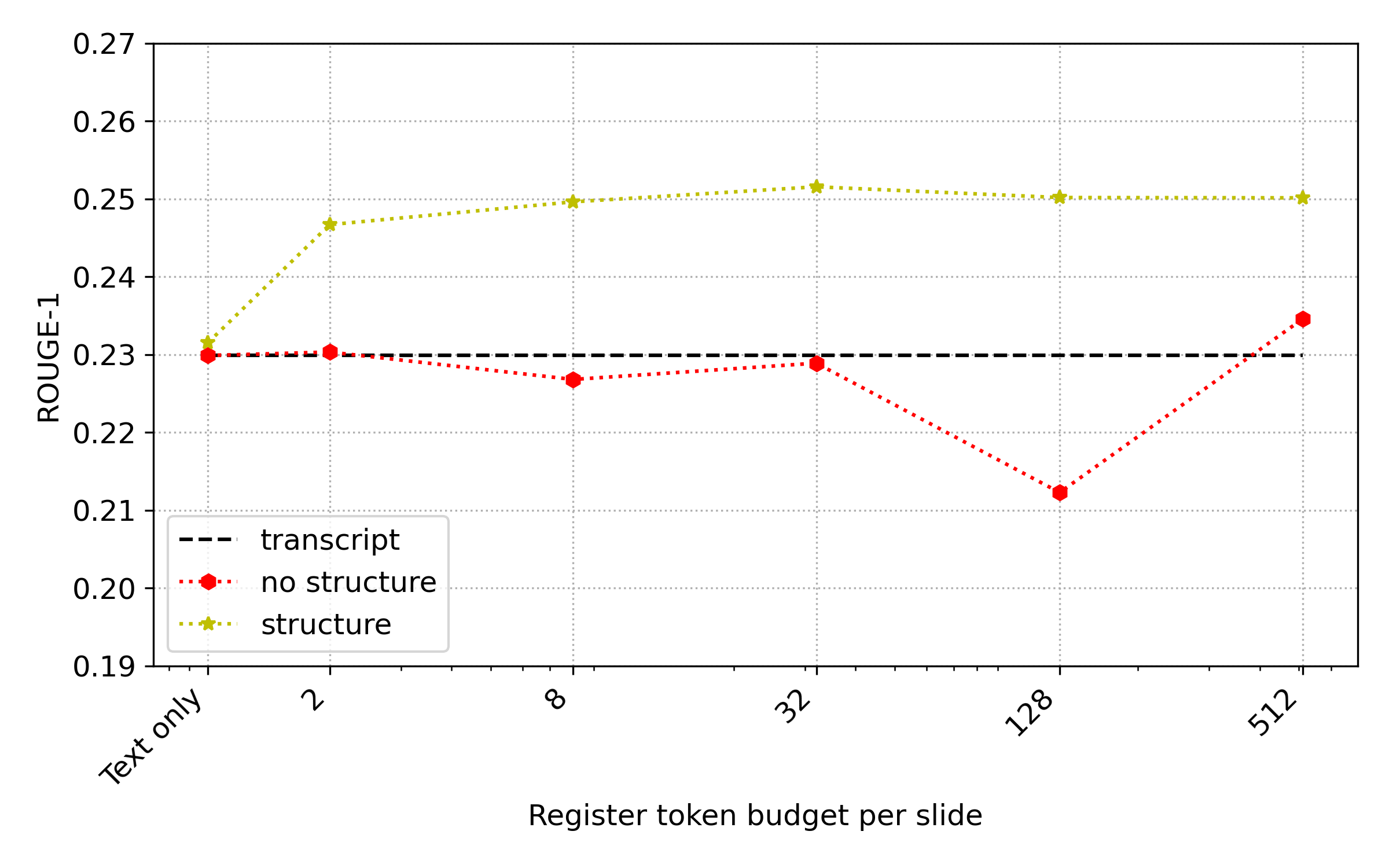}
    \caption{The addition of structure improves the Rouge score compared to the transcript alone}
    \label{fig:register-scaling}
\end{figure}

We conjecture that the interleaved format gives useful information about the structure of the presentation to the model. In order to verify and quantify this claim, we feed the model interleaved or concatenated modalities and evaluate the generated summaries.
We call these experiments \textit{structure} and \textit{no structure} respectively in Figure \ref{fig:register-scaling}.


Due to the scaling of the Transformer computation with sequence length \cite{vaswani_attention_2017}, a larger visual token budget mechanically increases the computation for a given document. The additional tokens can also be repurposed by the model for internal computations \cite{darcet_vision_2023}, in a phenomenon called "register" tokens. In an attempt to decouple this additional computation with the benefit of visual information, we conducted an ablation study by replacing the slides with placeholder black images. When slides are not included in the input, these placeholder slides are added to keep the same total computation budget, and watch for an effect similar to the "register" tokens discussed by \citet{darcet_vision_2023}.

The resulting experiments enable us to watch for the outcome of the structure information for summarization, decoupled from the effect of additional computations.
As illustrated in Figure \ref{fig:register-scaling}, scaling the register token budget for the placeholder slides does not show a trend on the measured Rouge score, but random variation around the mean that we believe is an artifact of the added noise.
However, the structure information offers a significant improvement compared to the text-only baseline.

\subsection{Model and visual token budget scaling}


\label{sec:visual_scaling}

\begin{figure}[h]
    \includegraphics[width=0.7\linewidth]{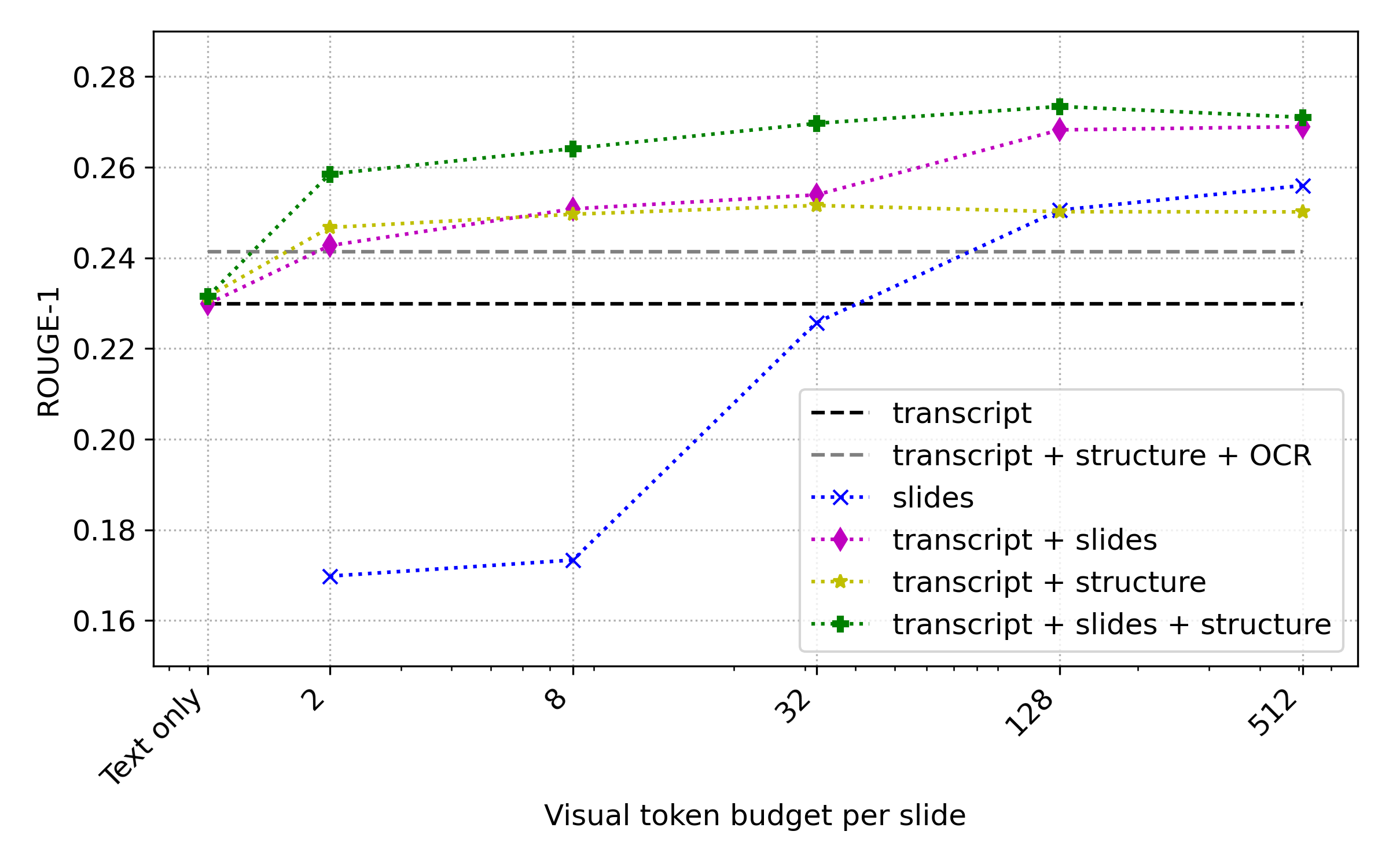}
    \caption{Rouge score with different visual token budgets}
    \label{fig:visual-scaling}
\end{figure}


Figure \ref{fig:visual-scaling} shows the scaling of the Rouge score with varying visual token budgets for various input representations.
With low visual token budgets, visual information is not as important as structure or text information. However, when the budget exceeds 32 tokens per slide, the visual information becomes the predominant modality.
At low budgets, the structure information vastly increases the results, but its importance lessens with more budget, to the point where it seems to converge with the unstructured text and image input at 512 tokens per slide.


Comparing the transcript only with the OCR-augmented textual baseline in Figure \ref{fig:visual-scaling}, we see that access to the text printed on the slide improves the performance. However, the slides can be more informative than the text printed on them. For example, the formatting of the text, the colors, and the illustrations can convey meaning.
In fact, the summaries generated with the visual input greatly exceed the textual baseline augmented with OCR, suggesting that the textual content of the slides covers only a small amount of the overall semantics they provide.


\begin{figure}[h]
    \includegraphics[width=0.7\linewidth]{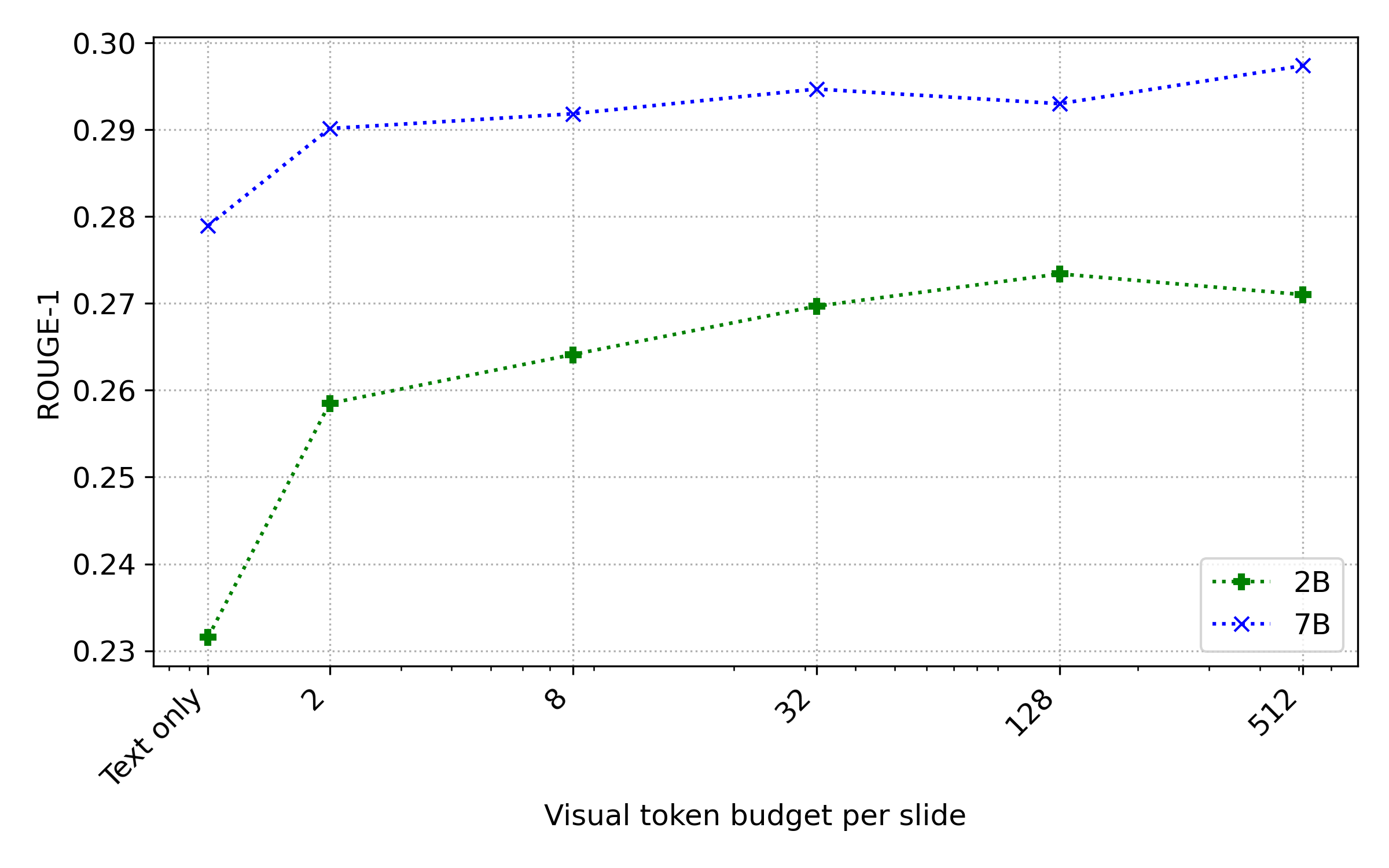}
    \caption{Rouge score with varying visual token budgets for different model sizes}
    \label{fig:model-scaling}
\end{figure}

As illustrated in Figure \ref{fig:model-scaling}, the improvement related to the scaling of the model from 2 billion to 7 billion parameters exceeds the increase caused by the additional visual and structure information. The larger model still benefits from the additional structure and modality.


\begin{figure}[h]
    \centering
    \includegraphics[width=0.7\linewidth]{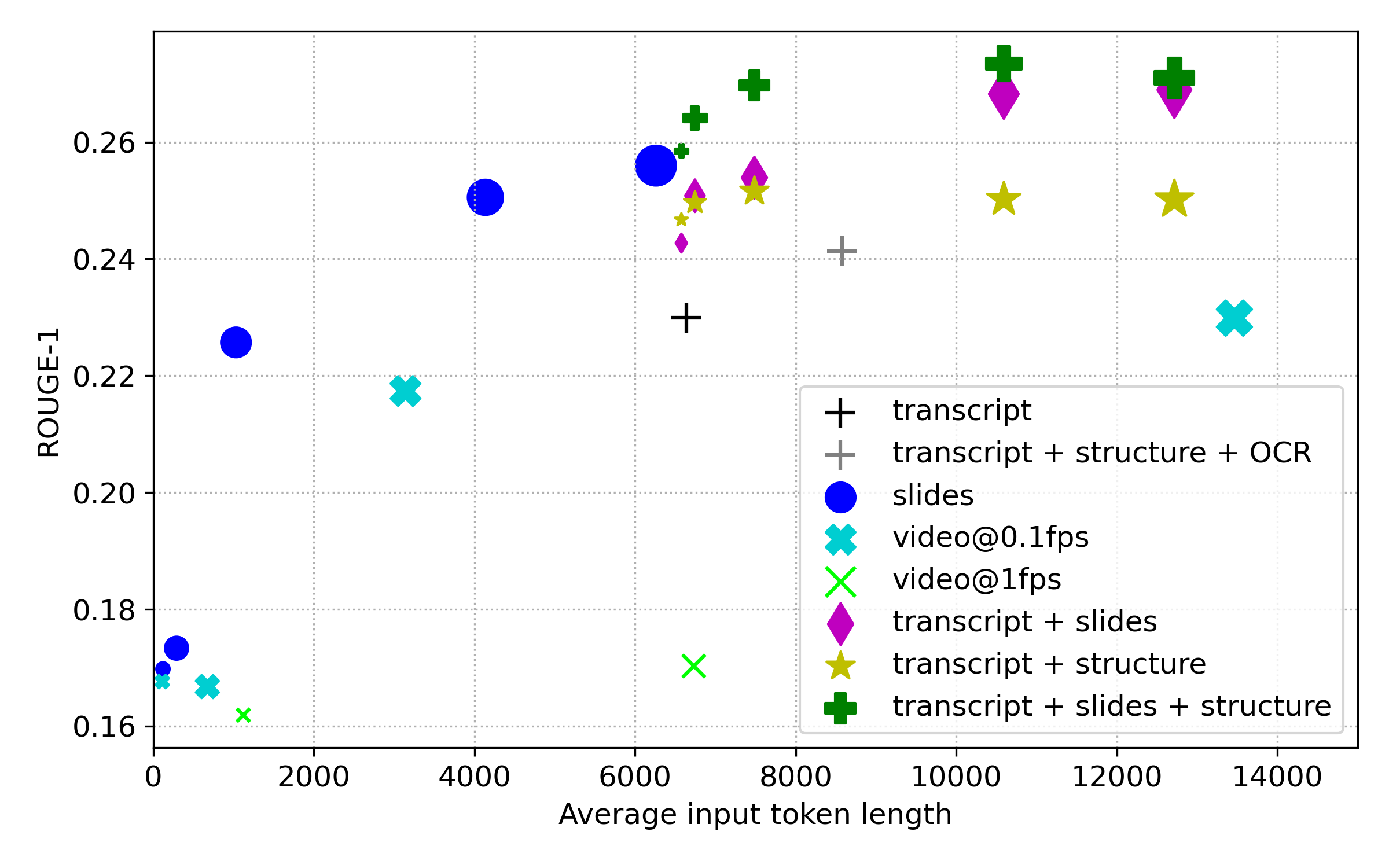}
    \caption{
    Slides-only and structured multimodal representations constitute input-length-optimal representations for different input lengths. Bigger icons depict higher visual token budget.}
    \label{fig:avg-length}
\end{figure}

\subsection{Cost-effective input settings}

The unimodal and multimodal input settings can have very different input lengths. In order to devise input-length-effective settings, we represent the Rouge score against the average input token length in Figure \ref{fig:avg-length}.
If we only account the cost of input size with the goal of improving Rouge score, the strategies resulting in the best tradeoffs draw a set of the input-length-optimal solutions, called Pareto frontier.

With an average input token length shorter than 6k, using only the slides as input is the most efficient strategy. For average token lengths larger than 6k, structured multimodal input with interleaved slides and transcript is the most effective.
These two strategies draw the Pareto frontier of the most efficient strategies with any input token length.
The text-only solutions that use transcript and/or OCR are significantly below this Pareto frontier, as are the solutions that use the whole video.

\subsection{Extractiveness and relevance of the extractive fragments with respect to input modalities}

\begin{figure}[h]
    \begin{subfigure}{0.45\linewidth}
        \includegraphics[width=\linewidth]{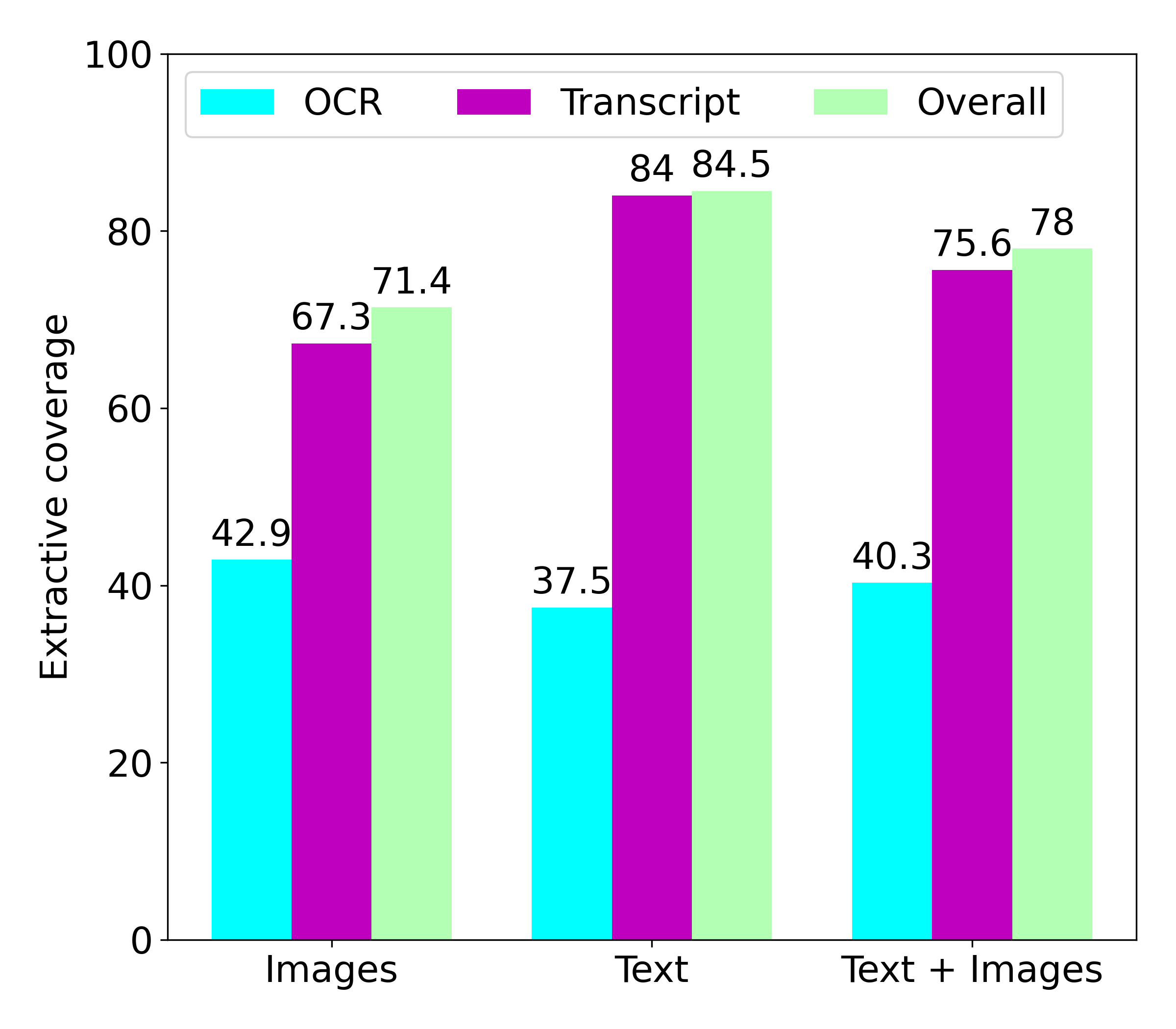}
        \caption{Unimodal or multimodal input}
        \label{fig:ext-coverage-modality}
    \end{subfigure}\hfil
    \begin{subfigure}{0.45\linewidth}
        \includegraphics[width=\linewidth]{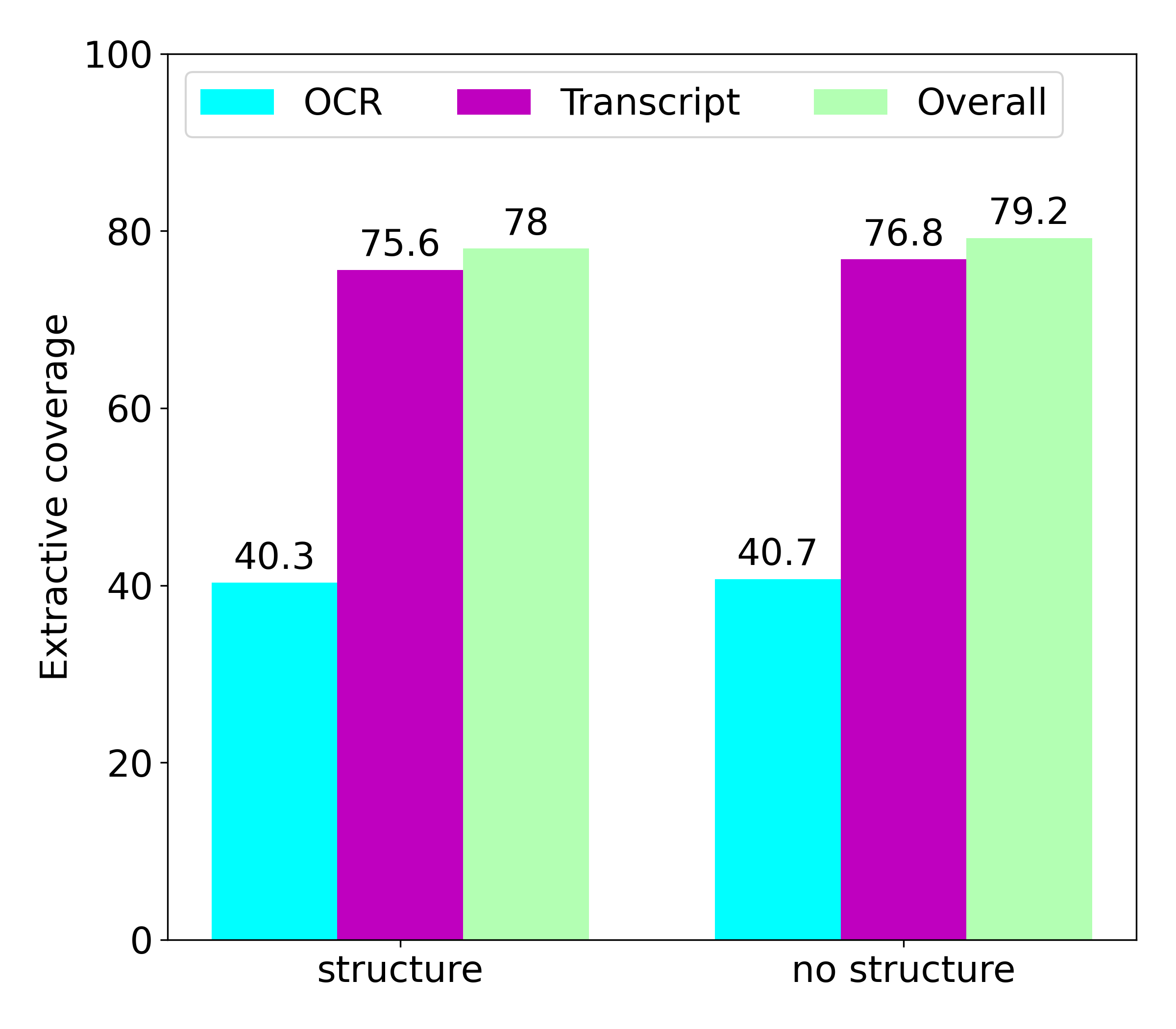}
        \caption{Structured or unstructured input}
        \label{fig:ext-coverage-structure}
    \end{subfigure}
    \caption{Extractive coverage of the predicted summaries with reference to input modality}
    \label{fig:ext-coverage}
\end{figure}

\begin{figure}[h]
    \begin{subfigure}{0.45\linewidth}
        \includegraphics[width=\linewidth]{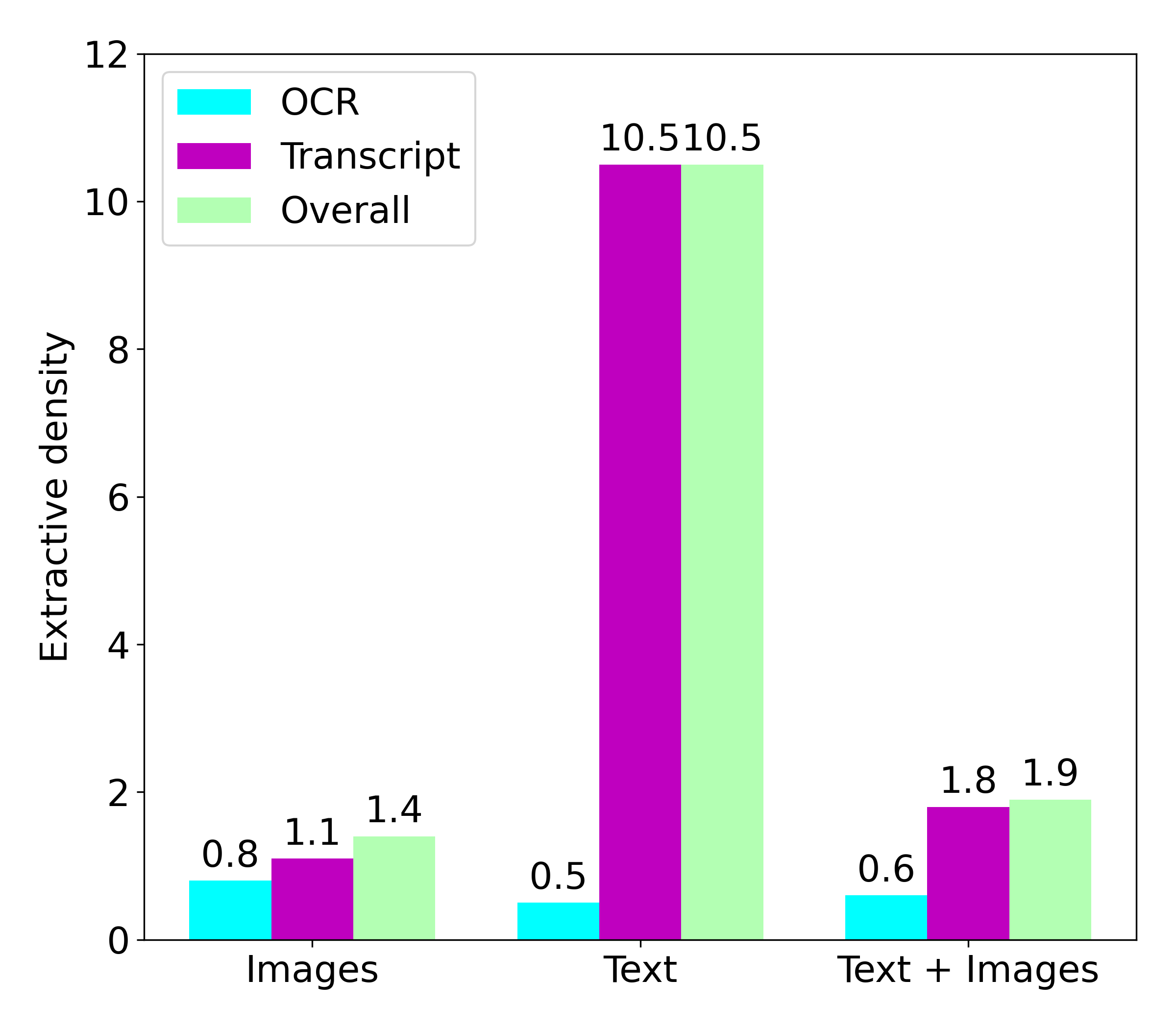}
        \caption{Unimodal or multimodal input}
        \label{fig:ext-density-modality}
    \end{subfigure}\hfil
    \begin{subfigure}{0.45\linewidth}
        \includegraphics[width=\linewidth]{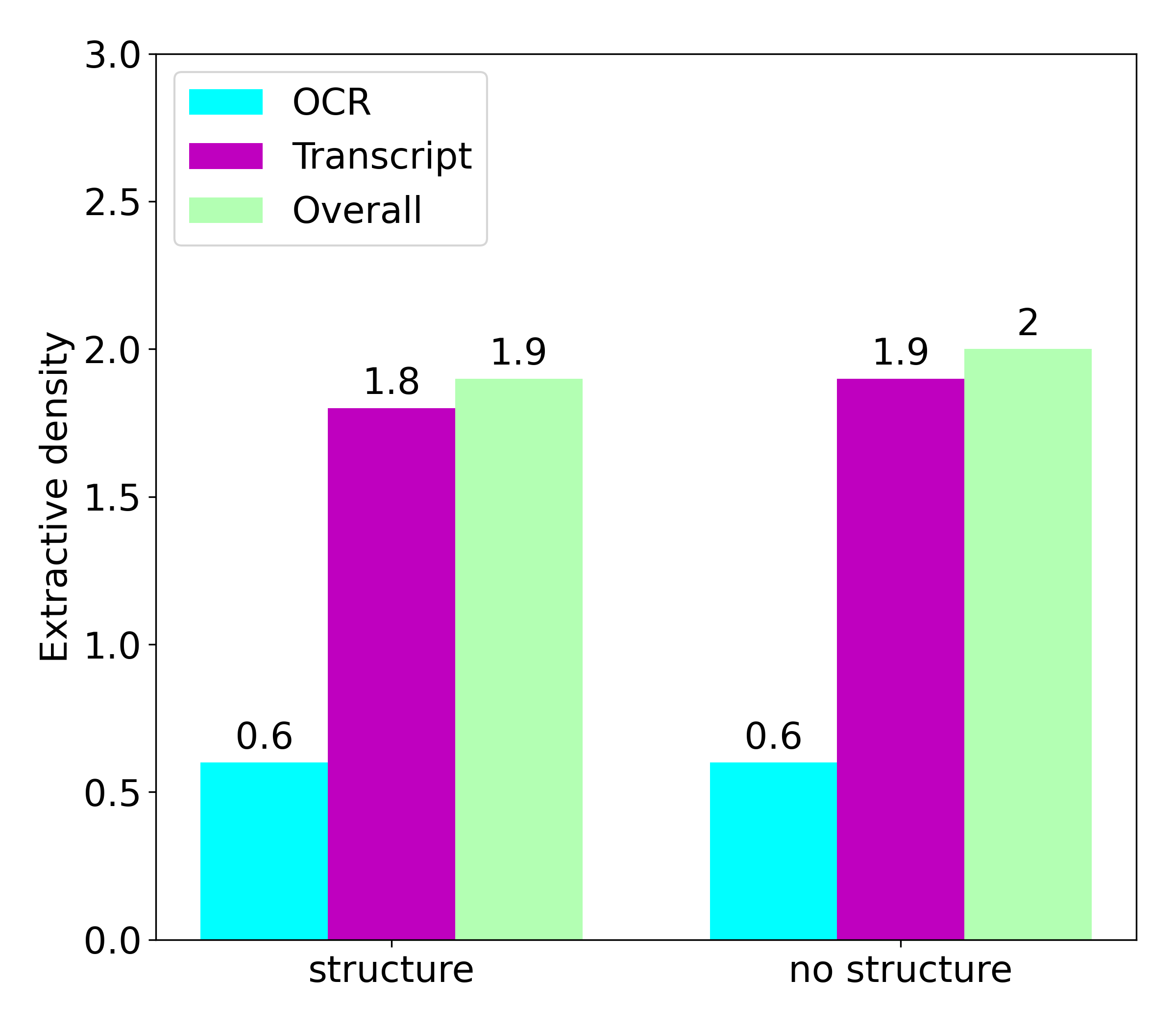}
        \caption{Structured or unstructured input}
        \label{fig:ext-density-structure}
    \end{subfigure}
    \caption{Extractive density of the predicted summaries with reference to input modality}
    \label{fig:ext-density}
\end{figure}

Figures \ref{fig:ext-coverage-modality} and \ref{fig:ext-density-modality} show the extractive coverage and density in unimodal and multimodal input settings, with respect to either the text extracted from the slides, from the speech or their concatenation, called \texttt{ocr}, \texttt{transcript} and \texttt{overall} modes respectively.
Compared to single-modality input, adding an additional modality results in less extractive summaries with respect to the original modality. Note that the \texttt{overall} mode is significantly biased towards the text extracted from speech, because of the imbalance illustrated in Figure \ref{fig:venn-token-count}, and as a consequence its change will be mostly correlated to the \texttt{transcript} mode.

\begin{figure}[h]
    \begin{subfigure}{0.45\linewidth}
        \includegraphics[width=\linewidth]{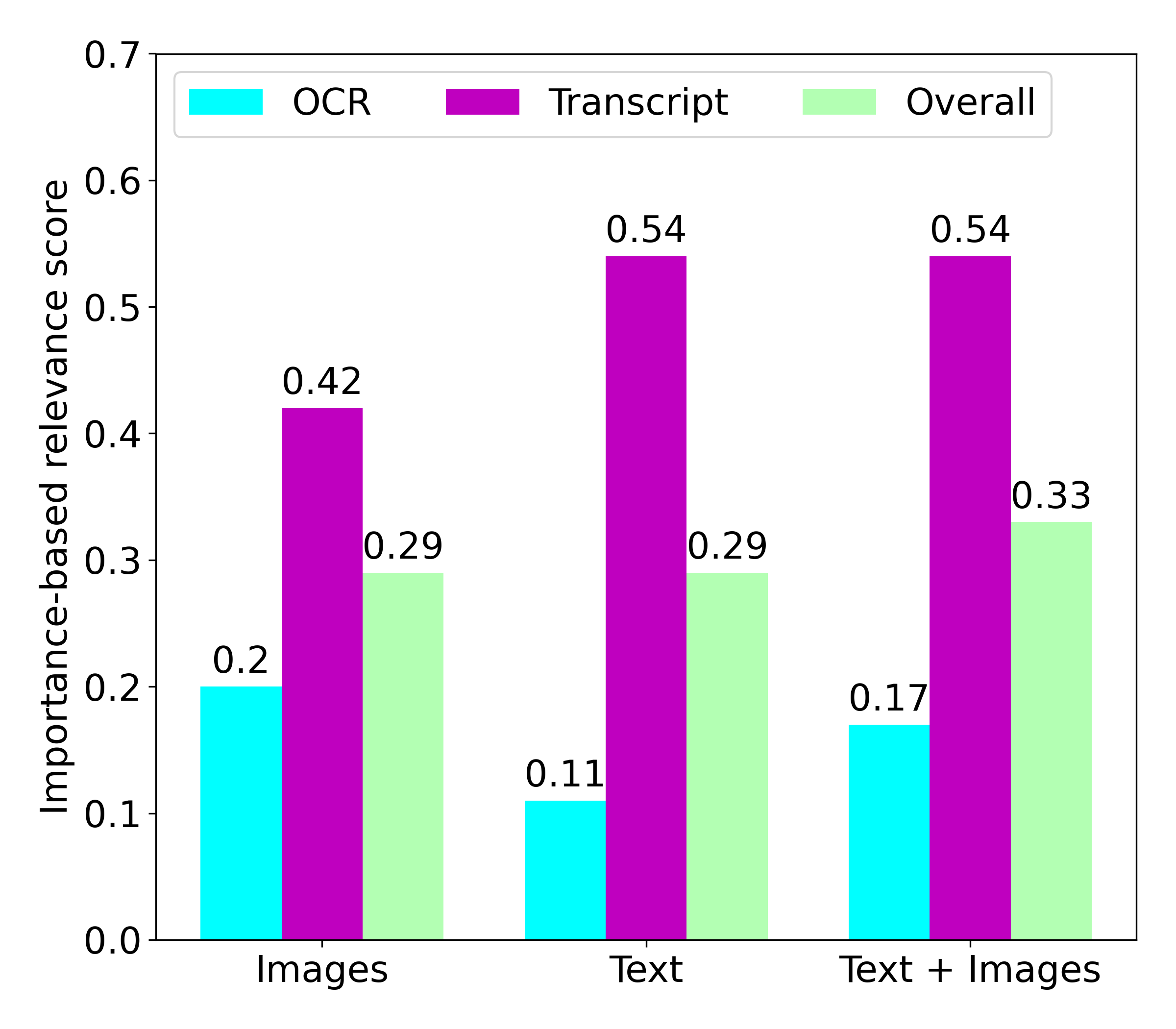}
        \caption{Unimodal or multimodal input}
        \label{fig:ext-importance-modality}
    \end{subfigure}\hfil
    \begin{subfigure}{0.45\linewidth}
        \includegraphics[width=\linewidth]{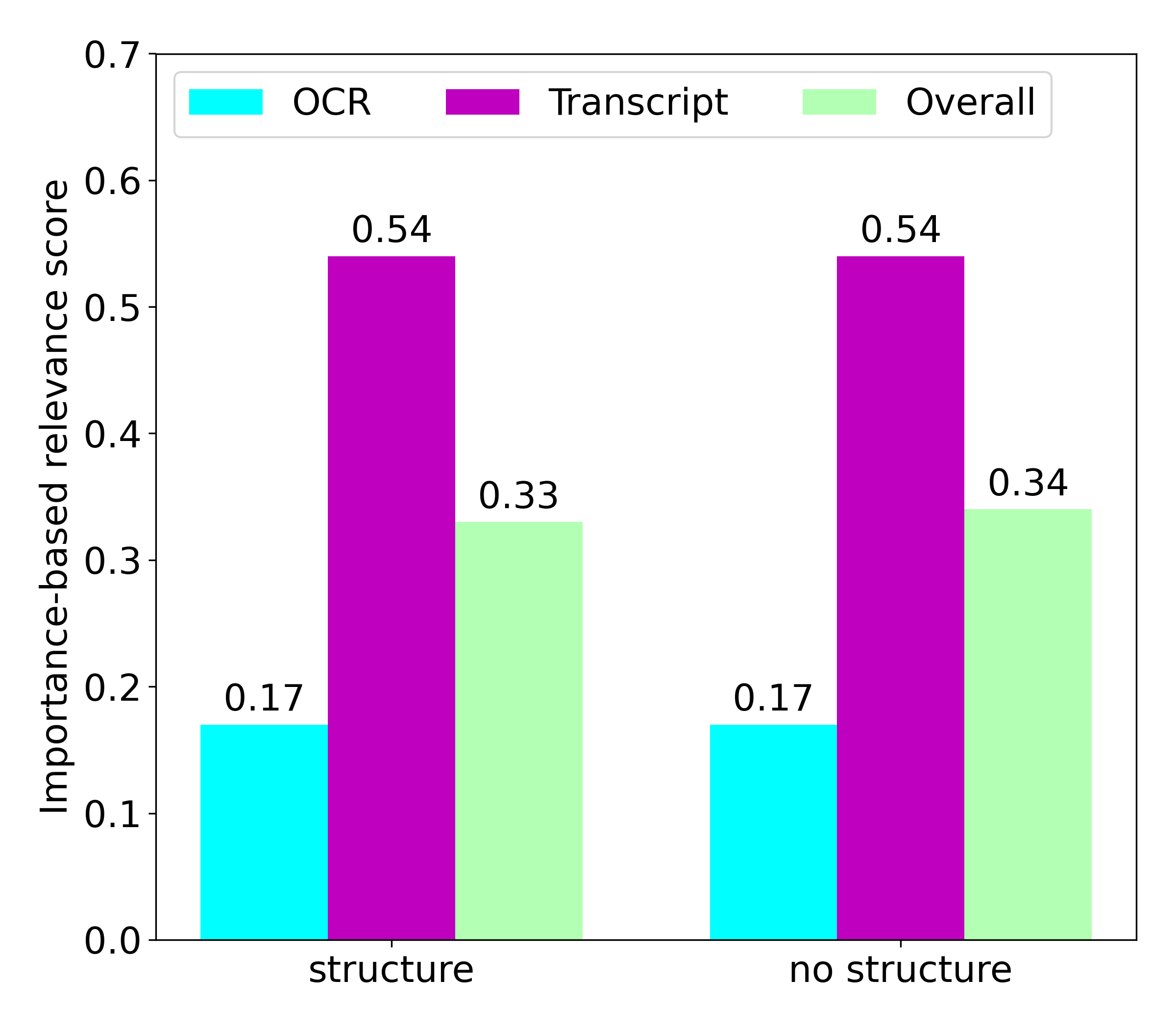}
        \caption{Structured or unstructured input}
        \label{fig:ext-importance-structure}
    \end{subfigure}
    \caption{Importance-based relevance score of the predicted summaries with reference to input modality}
    \label{fig:ext-importance}
\end{figure}

Figure \ref{fig:ext-importance-modality} shows the importance-based relevance score in unimodal and multimodal settings. Similar scores are observed in the \texttt{ocr} and \texttt{transcript} modes between the unimodal or multimodal inputs, although the summaries were shown to be less extractive. This score computes an overlap of $n$-grams between the summary and the source, weighted by the importance of the $n$-grams. A less extractive summary will mechanically have a lower score. Therefore, the extractive fragments in the multimodal setting are more relevant on average compared to the single-modality settings.



\begin{figure}[h]
    \begin{subfigure}{0.45\linewidth}
        \includegraphics[width=\linewidth]{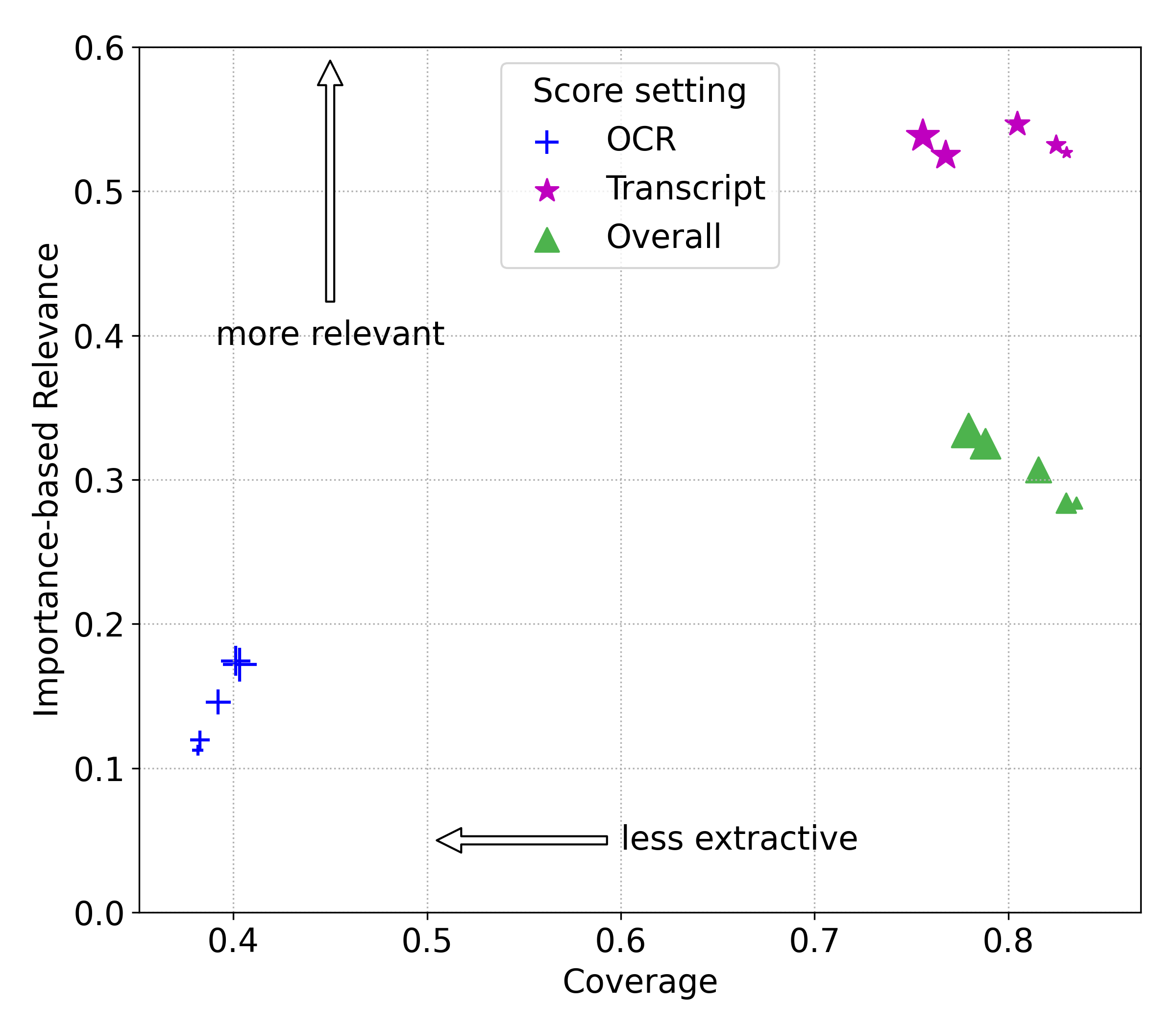}
        \caption{Coverage}
    \end{subfigure}\hfil
    \begin{subfigure}{0.45\linewidth}
        \includegraphics[width=\linewidth]{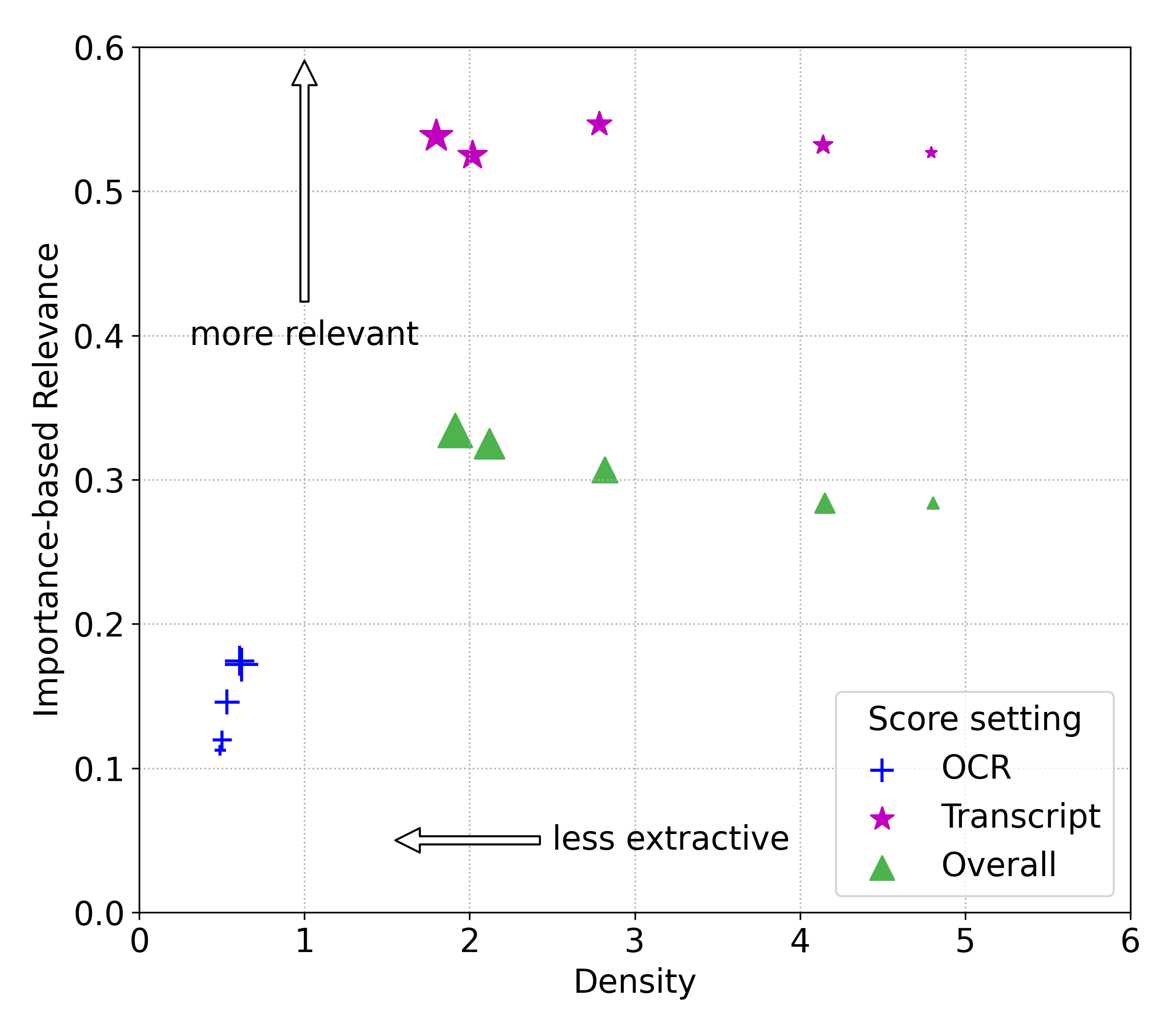}
        \caption{Density}
    \end{subfigure}
    \caption{Scaling the visual token budget balances the extractive source from transcript to OCR and improves the relevance with respect to the OCR and the overall document. Bigger icons depict higher visual token budgets.}
    \label{fig:relevance-extract}
\end{figure}

Figure \ref{fig:relevance-extract} illustrates this change in model behavior when scaling the visual token budget in the structured multimodal representation. The summaries become less extractive with respect to the speech (via transcript), while keeping a similar relevance, meaning that the extractive fragments are more relevant on average. At the same time, the summaries become more extractive and more relevant with respect to visual content (via OCR). Overall, the summaries are less extractive and more relevant.



We observe that with sufficient visual token budget, the structure and unstructured multimodal input seemed to converge to comparable performance. The extractive statistics and the importance metric pictured in Figures \ref{fig:ext-coverage-structure}, \ref{fig:ext-density-structure}, and \ref{fig:ext-importance-structure}, corroborate these findings. With large visual token budget, the model using structured or unstructured input behaves similarly with respect to the extractiveness and the importance of the extractive fragments. 

\subsection{Qualitative Analysis}

During manual inspection of the generated summaries, we notice different behaviors with different input settings for the model. For instance, summaries generated with a multimodal input tend to be more comprehensive than those generated with slides or transcripts alone.

As illustrated in Table \ref{tab:ex-spellcheck}, the visual information allows the model to correct misspelled words from the transcript, often proper nouns. With the example from Table \ref{tab:ex-context-slides} we see that it also includes contextual information that is not always found in the transcript, such as the name of the speaker and the venue and event where the presentation is taking place. 
Conversely, the speech tends to provide more detailed information compared to the slides, as displayed in Table \ref{tab:ex-context-transcript}.
The multimodal input seems to frequently trigger an unintended "bullet point" mode that was observed less frequently with unimodal inputs. This behaviour is shown in Table \ref{tab:ex-unprompted}.

\begin{table}[h]
    \tiny
    \centering
    \caption{Summaries generated from the presentation "\hl{Clojure}: a language for problem solvers", display the spell-checking abilities of the additional modalities}
    \begin{tabular}{p{0.3\linewidth}p{0.3\linewidth}p{0.3\linewidth}}
    \hline
        Summary with transcript input & Summary with slides input & Summary with multimodal input \\
        \hline
        The presentation by Stuart Halloway is about \hl{closure}, a programming language for problem solvers. The speaker introduces the topic of \hl{closure} and its importance in software development. He discusses the characteristics of \hl{closure}, such as being knowledgeable, flexible, powerful, and smart. He also mentions the challenges of working with transient data structures and the need for persistent data structures. Halloway explains how \hl{closure} uses persistent data structures to solve these challenges and how it provides direct access to the platform. He also discusses the benefits of using \hl{closure}, such as being able to build systems quickly and having a powerful set of persistent data structures. The presentation ends with a summary of the benefits of using \hl{closure} and a call to action for attendees to learn more about the language. & The presentation is about \hl{Clojure}, a language designed for problem solvers. The speaker, Stuart Halloway, introduces \hl{Clojure} and its features, such as its ability to handle transience and persistence, and its use in various programming contexts. The presentation includes slides that explain the language's syntax and features, as well as examples of how it can be used to solve problems. The slides also show how \hl{Clojure} can be used to implement interfaces and work with collections, directories, files, XML, JSON, and ResultSets. The presentation ends with a question and answer session and a list of resources for further learning. & The presentation by Stuart Halloway is about \hl{Clojure}, a programming language for problem solvers. The main focus of the presentation is to explain the benefits of using \hl{Clojure} for software development, particularly in terms of flexibility, power, and knowledge. Halloway discusses the importance of having knowledge in software development and how \hl{Clojure} provides a way to build systems that are knowledgeable and powerful. He also emphasizes the importance of having flexible software systems that can adapt to changing requirements. The presentation includes slides that explain the different types of data structures in \hl{Clojure}, such as vectors, maps, sets, and lists, and how they can be used to build powerful and flexible software systems. Halloway also discusses the benefits of using \hl{Clojure} for developing large-scale systems, such as the ability to build systems that are composable and can be easily modified and extended. The presentation also includes a demonstration of \hl{Clojure} in action, using a code example to show how it can be used to build a simple program. Overall, the presentation provides a comprehensive overview of \hl{Clojure} and its benefits for software development.
        \\
        \hline
    \end{tabular}
    \label{tab:ex-spellcheck}
\end{table}

\begin{table}[h]
    \tiny
    \centering
    \caption{Summaries generated from the presentation "Keyword Args — the killer Ruby feature you aren't using", illustrating the additional context when leveraging the slides}
    \begin{tabular}{p{0.3\linewidth}p{0.3\linewidth}p{0.3\linewidth}}
    \hline
        Summary with transcript input & Summary with slides input & Summary with multimodal input \\
        \hline
        The speaker is introducing the topic of keyword arguments in Ruby programming. They explain that keyword arguments are a feature in Ruby that allows for more flexible and functional programming. They provide examples of how keyword arguments can be used to pass arguments to functions and how they can be used to make functions more flexible. The speaker also discusses the benefits of using keyword arguments over positional arguments, such as making functions more clear and flexible. & The presentation is about \hl{RubyConf Cincinnati 2016, featuring a talk by Guyren G Howe titled "Keyword Args - the ikller Ruby feature you aren't using."} The speaker discusses the use of keyword arguments in Ruby, highlighting their benefits such as making code more intuitive, clearer, more flexible, and more functional. The presentation includes slides that explain the different types of keyword arguments, such as positional, required, and optional arguments, and how they can be used to make code more expressive. The speaker also demonstrates how keyword arguments can be used to create more flexible and functional code. The presentation concludes with a discussion on dependency injection and context funneling, which are other Ruby features that are not commonly used. & The presentation by \hl{Guyren G Howe at RubyConf Cincinnati 2016} focuses on keyword arguments in Ruby programming. The speaker explains that keyword arguments are a feature in Ruby that allows for more flexible and functional programming. They are particularly useful for passing positional arguments, optional arguments, and keyword arguments to functions. The speaker provides examples of how keyword arguments can make code more clear, more flexible, and more functional. He also discusses the benefits of using keyword arguments over positional arguments, such as the ability to pass in optional arguments and the ability to override default values. The presentation concludes with a summary of the key points and a call to action for attendees to consider using keyword arguments in their Ruby code.
        \\
        \hline
    \end{tabular}
    \label{tab:ex-context-slides}
\end{table}

\begin{table}[h]
    \tiny
    \centering
    \caption{Summaries generated from the presentation "UrbanSim2: Simulating the Connected Metropolis", illustrating the additional context when leveraging the transcript: the slides mention \hl{Paul Waddell} but the speaker is \hl{Fletcher Foti} as said in the presentation}
    \begin{tabular}{p{0.3\linewidth}p{0.3\linewidth}p{0.3\linewidth}}
    \hline
        Summary with transcript input & Summary with slides input & Summary with multimodal input \\
        \hline
        \hl{Fletcher Foti} is a PhD in city planning who is now a chief data scientist at Synthesity. He is presenting on UrbanSim, a simulation tool that simulates the connected metropolis. He explains that UrbanSim is an agent-based simulation of regional real estate markets, and that it captures individual households, individual jobs, and the decisions that individual households make. He also mentions that UrbanSim is an open-source project that is BSD licensed. & The presentation is titled "What is UrbanSim?" and is part of the FOSS4G PDX 2014 conference. The speaker, \hl{Paul Waddell}, is a professor at the University of California, Berkeley. The presentation covers the purpose of UrbanSim, which is an agent-based simulation of regional real estate markets. The simulation is written in Java in 2000 and Python in 2006. The code is implemented to use Pandas this year. The presentation also discusses the regional task, which includes reducing greenhouse gas emissions, housing the region's population at all income levels, and embodying local visions. The speaker also mentions the Open Source Stack, which includes tools like Pandas, Python, and Scikit. & The presentation by \hl{Fletcher Foti} focused on UrbanSim, a statistical model used for simulating the connected metropolis. The speaker explained that UrbanSim is used to forecast change in population, employment, and the built environment over a 30-year planning horizon for a given metropolitan region. The model is agent-based, simulating individual households, jobs, and land use decisions. The speaker also highlighted the open-source nature of the project, with UrbanSim being written in Java in 2000 and Python in 2006. The model was re-implemented in 2013 on new scientific Python tools, specifically Pandas. The speaker also discussed the open-source stack, which includes Python, Pandas, and other libraries. The presentation included a demo of the Urban Data Science Toolkit, which includes tools for network analysis, data visualization, and web mapping. The speaker also mentioned the Sync City project, which is a large-scale transportation model required by federal law.
        \\
        \hline
    \end{tabular}
    \label{tab:ex-context-transcript}
\end{table}

\begin{table}[h]
    \tiny
    \centering
    \caption{Summaries generated from the presentation "Genode meets the Pinephone", exemplifying the unprompted use of bullet point lists with the multimodal input}
    \begin{tabular}{p{0.3\linewidth}p{0.3\linewidth}p{0.3\linewidth}}
    \hline
        Summary with transcript input & Summary with slides input & Summary with multimodal input \\
        \hline
        The talk was about the last year's journey to bring together the GenoOS framework and the Pinephone. The speaker, Norman Feske, from GenoDepps, the developer behind the GenoOS framework, gave a presentation on the topic. & The presentation is titled "Genode meets the Pinephone" and is given by Norman Feske. The presentation covers the motivation behind the development of Genode, which is a microkernel-based operating system. The presentation also discusses the corporate motives of the developers and the sentiments shared by members of civil society. The development story of Genode is presented, including the boot steps, device access, and cascaded authorities. The presentation also covers the development of Genode on the Pinephone, including the device tree, device drivers, and the Pinephone as a development platform. The presentation ends with a preview of the Genode OS on the Pinephone and a list of goals for 2022. & The presentation by Norman Feske at FOSDEM 2022 focuses on the development of GenoOS, a microkernel-based operating system for smartphones. The presentation covers the motivation behind the project, the development process, and the current status of the project. \hl{Key points include:
- The project aims to combine the GenoOS framework with the PinePhone, a smartphone that uses a custom microkernel.
- The development process involves booting the device, creating a custom Linux kernel, and porting the kernel to the PinePhone.
- The development team has made progress in porting the kernel and has successfully run a small application on the PinePhone.
- The team is working on improving performance and adding new features, such as persistent storage and mobile data connectivity.
- The presentation also includes a demonstration of running a small application on the PinePhone and a preview of the Sculpt OS on the PinePhone.
- The team plans to continue working on the project in 2022, with a focus on video telephony, persistent storage, mobile data connectivity, power management, and performance improvements.}
        \\
        \hline
    \end{tabular}
    \label{tab:ex-unprompted}
\end{table}


\section{Discussion}

We think that the behaviors observed in the qualitative analysis are a consequence of the training data domain of VLMs. Specifically, the underrepresentation of conflict and mismatch between the textual and visual input.

Qwen2-VL, similarly to most VLMs, is trained with data covering multiple tasks that rely mostly on translating the information from one modality to another, \textit{eg} captioning or visual question answering. \citet{mckinzie_mm1_2025} observed that common VLM evaluation is focused on captioning problems, with 3 out of 8 benchmarks in their study being captioning. They discriminate the multimodal data used to train VLMs between captioning image-text pairs and interleaved image-text documents. They describe their interleaved image-text data as longer than the captioning data and with more diverse text with less relevance on average to images.

We believe that the interleaved image-text that is extracted from multimodal presentations describes a different flavor of cross-modal interactions compared to the interleaved image-text documents typically found in VLMs' training data.
The latter's interleaved image-text documents mainly come from articles, in which the images are usually used for merely illustrative purposes.

\subsection{Non redundancy and conflict between modalities}

In multimodal presentations, information can be split between modalities with limited redundancy, as illustrated in Tables \ref{tab:ex-context-slides} and \ref{tab:ex-context-transcript}. In order to create a comprehensive summary, a VLM should be able to cross-reference modalities.
Some visual question answering benchmarks \cite{lerner_viquae_2022, chen_can_2023} are designed to require cross-referencing between modalities. They show the limited performance of VLMs in this task and their dependence on external tools such as information retrieval pipelines.


Additionally, mismatched or conflicting information can be presented across modalities. For example, when the speaker verbally states a fact that contradicts the content displayed on a slide, due to a typographical error, a last-minute change (\textit{cf} Table \ref{tab:ex-context-transcript}), or an \textit{erratum}. The contradiction can also be sourced from inaccuracies during preprocessing, \textit{eg} the transcription (\textit{cf} Table \ref{tab:ex-spellcheck}).
In our understanding of the training data domains, VLMs are not substantially faced with this type of data during training, and therefore are not explicitly trained to handle these situations.
\citet{liu_insight_2024} explored the conflict between a VLM's internal knowledge and its visual input. They proposed a prompting strategy as a way to steer the VLM into trusting vision over internal knowledge. Sarcasm detection in VLMs \cite{castro_towards_2019} is another line of work that represents a typical example of conflicting modalities. In these cases, the task implicitly defines a \textit{stronger} modality, on which trust should be focused. In multimodal presentations, errors can appear in both modalities, even in the same sample. Consequently, no overall \textit{stronger} modality can be chosen.
\citet{liu_learn_2018} described \textit{weak} modalities as modalities that are very noisy but could provide relevant information on a per sample basis. They advocate for a multiplicative method rather than additive to merge modalities in a late-fusion scenario, effectively leveraging the lower uncertainty, or entropy, associated with weak modalities in relevant situations. Modern VLMs are early-fusion models. Therefore it is not trivial to adapt this uncertainty-based method to merge decisions based on each modality.

\subsection{Visual and written concepts entanglement in the visual modality}

The visual modality contains information that can be depicted or written.
\citet{goh_multimodal_2021} showed that printing textual information over the input on multimodal vision models can steer the output, acting as "typographical attacks". We believe that the unprompted "bullet point" mode (\textit{cf} Table \ref{tab:ex-unprompted}) which follows the format usually seen in slides, emerging in multimodal contexts, could derive from this phenomenon. \citet{materzynska_disentangling_2022} further studied the entanglement of visual and written concepts in CLIP. They hypothesized that it is a consequence of the prevalence of text next to the concept they represent in the training data, such as signs and labels.
Overall, text written in the visual modality, characteristic of the text-heavy visual stream of multimodal presentations, has an unpredictable effect on the model behavior. It can be used as a way to interfere with the instructions via typographical attacks or be confused with visual concepts.

\subsection{VLMs are mostly not agnostic to the input modalities}

Similar information can be provided to VLMs through visual or textual modalities, in practice resulting in very different outcomes.
\citet{liang_mind_2022} observed a modality gap in the embeddings of vision-language encoders such as CLIP. This gap is characterized by the representation of an image being closer to representations of different images than to the representation of its associated caption in the shared representation space.
\citet{zhu_unraveling_2024} observed conflicts in the knowledge embedded in the model parameters depending on the input modality, even when the model accurately captions the images.
Alternatively, \citet{templeton_scaling_2024} managed to extract monosemantic features from a VLM, some of which were multimodal and multilingual. These features are directions in the activation space that represent concepts that are agnostic to language and modality. \citet{zhang_cross-modal_2024} described a phenomenon dubbed cross-modal consistency associated with the performance gap of a VLM on the same task depending on the input modality. In their experiments, this gap is substantial and results in the VLM being inconsistent with different input modalities. \citet{deng_words_2025} observed a "Blind Faith in Text" effect where VLMs tend to trust textual input when it is inconsistent with the visual data, or incorrect. In contrast, in our experiments, the model was able to correct errors in the transcript by relying on visual information (\textit{cf} Table \ref{tab:ex-spellcheck}), probably because it was informed in its prompt that the text is a transcript, therefore implicitly error prone.

These works illustrate some of the inherent challenges in the abstractive summarization of multimodal presentations. The internal representation of VLMs can contain modality-agnostic features but is still overall vastly inconsistent between and inside modalities.



\section{Conclusion}

Available state-of-the-art open-weights Vision-Language Models are still overall quite unsuccessful at summarizing long multimodal presentations. We find some of them, notably Qwen2-VL, which was likely trained on adjacent tasks that involve long-range text-heavy interleaved image-text documents, to express promising downstream performance for multimodal summarization.

Through quantitative and qualitative analyses, we show the benefit of representing the input presentation record as interleaved slides and transcript. This structured multimodal representation results in higher scores for VLMs and extends the Pareto frontier of input-length-optimal methods beyond the use of slides alone.
The use of both modalities offers many benefits, such as providing additional information and enabling spell checking through cross-referencing modalities. Multimodal systems produce summaries that are less extractive with respect to speech or text displayed on slides. They also include more relevant elements of the presentation, sourced from either one or both modalities. However, discrepancies between transcript and slides pose challenges for the VLM and occasionally exhibit impromptu side effects.

We discuss some of the challenges that arise from interactions between conflicting modalities in multimodal presentations. Consequently, we advocate for the inclusion of more diverse cross-modality interactions in the training data mixtures for VLMs, specifically documents involving mismatched or conflicting modalities. Such data is commonly encountered in real-world scenarios such as multimodal presentations, where the visual content displayed is often commented upon, expanded, or even criticized in the accompanying speech. Data of this nature can also be sourced from reviews and commentaries, or created synthetically by altering a modality \cite{deng_words_2025}.

By incorporating more documents with conflicting modalities into datasets and benchmarks, future research can better understand and mitigate the effects of cross-modal interactions in VLMs. This would pave the way for the development of more robust, reliable, and trustworthy multimodal systems.

\bibliographystyle{ACM-Reference-Format}
\bibliography{vlm-summarization}


\begin{thebibliography}{51}


\ifx \showCODEN    \undefined \def \showCODEN     #1{\unskip}     \fi
\ifx \showISBNx    \undefined \def \showISBNx     #1{\unskip}     \fi
\ifx \showISBNxiii \undefined \def \showISBNxiii  #1{\unskip}     \fi
\ifx \showISSN     \undefined \def \showISSN      #1{\unskip}     \fi
\ifx \showLCCN     \undefined \def \showLCCN      #1{\unskip}     \fi
\ifx \shownote     \undefined \def \shownote      #1{#1}          \fi
\ifx \showarticletitle \undefined \def \showarticletitle #1{#1}   \fi
\ifx \showURL      \undefined \def \showURL       {\relax}        \fi
\providecommand\bibfield[2]{#2}
\providecommand\bibinfo[2]{#2}
\providecommand\natexlab[1]{#1}
\providecommand\showeprint[2][]{arXiv:#2}

\bibitem[Abdin et~al\mbox{.}(2024)]%
        {abdin_phi-3_2024}
\bibfield{author}{\bibinfo{person}{Marah~I. Abdin}, \bibinfo{person}{Sam~Ade Jacobs}, \bibinfo{person}{Ammar~Ahmad Awan}, \bibinfo{person}{Jyoti Aneja}, \bibinfo{person}{Ahmed Awadallah}, \bibinfo{person}{Hany Awadalla}, \bibinfo{person}{Nguyen Bach}, \bibinfo{person}{Amit Bahree}, \bibinfo{person}{Arash Bakhtiari}, \bibinfo{person}{Harkirat~S. Behl}, \bibinfo{person}{Alon Benhaim}, \bibinfo{person}{Misha Bilenko}, \bibinfo{person}{Johan Bjorck}, \bibinfo{person}{Sébastien Bubeck}, \bibinfo{person}{Martin Cai}, \bibinfo{person}{Caio César~Teodoro Mendes}, \bibinfo{person}{Weizhu Chen}, \bibinfo{person}{Vishrav Chaudhary}, \bibinfo{person}{Parul Chopra}, \bibinfo{person}{Allie~Del Giorno}, \bibinfo{person}{Gustavo~de Rosa}, \bibinfo{person}{Matthew Dixon}, \bibinfo{person}{Ronen Eldan}, \bibinfo{person}{Dan Iter}, \bibinfo{person}{Amit Garg}, \bibinfo{person}{Abhishek Goswami}, \bibinfo{person}{Suriya Gunasekar}, \bibinfo{person}{Emman Haider}, \bibinfo{person}{Junheng Hao}, \bibinfo{person}{Russell~J.
  Hewett}, \bibinfo{person}{Jamie Huynh}, \bibinfo{person}{Mojan Javaheripi}, \bibinfo{person}{Xin Jin}, \bibinfo{person}{Piero Kauffmann}, \bibinfo{person}{Nikos Karampatziakis}, \bibinfo{person}{Dongwoo Kim}, \bibinfo{person}{Mahoud Khademi}, \bibinfo{person}{Lev Kurilenko}, \bibinfo{person}{James~R. Lee}, \bibinfo{person}{Yin~Tat Lee}, \bibinfo{person}{Yuanzhi Li}, \bibinfo{person}{Chen Liang}, \bibinfo{person}{Weishung Liu}, \bibinfo{person}{Eric Lin}, \bibinfo{person}{Zeqi Lin}, \bibinfo{person}{Piyush Madan}, \bibinfo{person}{Arindam Mitra}, \bibinfo{person}{Hardik Modi}, \bibinfo{person}{Anh Nguyen}, \bibinfo{person}{Brandon Norick}, \bibinfo{person}{Barun Patra}, \bibinfo{person}{Daniel Perez-Becker}, \bibinfo{person}{Thomas Portet}, \bibinfo{person}{Reid Pryzant}, \bibinfo{person}{Heyang Qin}, \bibinfo{person}{Marko Radmilac}, \bibinfo{person}{Corby Rosset}, \bibinfo{person}{Sambudha Roy}, \bibinfo{person}{Olatunji Ruwase}, \bibinfo{person}{Olli Saarikivi}, \bibinfo{person}{Amin Saied},
  \bibinfo{person}{Adil Salim}, \bibinfo{person}{Michael Santacroce}, \bibinfo{person}{Shital Shah}, \bibinfo{person}{Ning Shang}, \bibinfo{person}{Hiteshi Sharma}, \bibinfo{person}{Xia Song}, \bibinfo{person}{Masahiro Tanaka}, \bibinfo{person}{Xin Wang}, \bibinfo{person}{Rachel Ward}, \bibinfo{person}{Guanhua Wang}, \bibinfo{person}{Philipp Witte}, \bibinfo{person}{Michael Wyatt}, \bibinfo{person}{Can Xu}, \bibinfo{person}{Jiahang Xu}, \bibinfo{person}{Sonali Yadav}, \bibinfo{person}{Fan Yang}, \bibinfo{person}{Ziyi Yang}, \bibinfo{person}{Donghan Yu}, \bibinfo{person}{Chengruidong Zhang}, \bibinfo{person}{Cyril Zhang}, \bibinfo{person}{Jianwen Zhang}, \bibinfo{person}{Li~Lyna Zhang}, \bibinfo{person}{Yi Zhang}, \bibinfo{person}{Yue Zhang}, \bibinfo{person}{Yunan Zhang}, {and} \bibinfo{person}{Xiren Zhou}.} \bibinfo{year}{2024}\natexlab{}.
\newblock \showarticletitle{Phi-3 {Technical} {Report}: {A} {Highly} {Capable} {Language} {Model} {Locally} on {Your} {Phone}}.
\newblock \bibinfo{journal}{\emph{CoRR}} (\bibinfo{date}{Jan.} \bibinfo{year}{2024}).
\newblock
\urldef\tempurl%
\url{https://openreview.net/forum?id=k05PGWgTv2}
\showURL{%
\tempurl}


\bibitem[Ainslie et~al\mbox{.}(2020)]%
        {ainslie_etc_2020}
\bibfield{author}{\bibinfo{person}{Joshua Ainslie}, \bibinfo{person}{Santiago Ontanon}, \bibinfo{person}{Chris Alberti}, \bibinfo{person}{Vaclav Cvicek}, \bibinfo{person}{Zachary Fisher}, \bibinfo{person}{Philip Pham}, \bibinfo{person}{Anirudh Ravula}, \bibinfo{person}{Sumit Sanghai}, \bibinfo{person}{Qifan Wang}, {and} \bibinfo{person}{Li Yang}.} \bibinfo{year}{2020}\natexlab{}.
\newblock \showarticletitle{{ETC}: {Encoding} {Long} and {Structured} {Inputs} in {Transformers}}. In \bibinfo{booktitle}{\emph{Proceedings of the 2020 {Conference} on {Empirical} {Methods} in {Natural} {Language} {Processing} ({EMNLP})}}, \bibfield{editor}{\bibinfo{person}{Bonnie Webber}, \bibinfo{person}{Trevor Cohn}, \bibinfo{person}{Yulan He}, {and} \bibinfo{person}{Yang Liu}} (Eds.). \bibinfo{publisher}{Association for Computational Linguistics}, \bibinfo{address}{Online}, \bibinfo{pages}{268--284}.
\newblock
\href{https://doi.org/10.18653/v1/2020.emnlp-main.19}{doi:\nolinkurl{10.18653/v1/2020.emnlp-main.19}}


\bibitem[Alaa et~al\mbox{.}(2024)]%
        {alaa_video_2024}
\bibfield{author}{\bibinfo{person}{Toqa Alaa}, \bibinfo{person}{Ahmad Mongy}, \bibinfo{person}{Assem Bakr}, \bibinfo{person}{Mariam Diab}, {and} \bibinfo{person}{Walid Gomaa}.} \bibinfo{year}{2024}\natexlab{}.
\newblock \bibinfo{title}{Video {Summarization} {Techniques}: {A} {Comprehensive} {Review}}.
\newblock
\href{https://doi.org/10.48550/arXiv.2410.04449}{doi:\nolinkurl{10.48550/arXiv.2410.04449}}
\newblock
\shownote{arXiv:2410.04449 [cs]}.


\bibitem[Alayrac et~al\mbox{.}(2022)]%
        {alayrac_flamingo_2022}
\bibfield{author}{\bibinfo{person}{Jean-Baptiste Alayrac}, \bibinfo{person}{Jeff Donahue}, \bibinfo{person}{Pauline Luc}, \bibinfo{person}{Antoine Miech}, \bibinfo{person}{Iain Barr}, \bibinfo{person}{Yana Hasson}, \bibinfo{person}{Karel Lenc}, \bibinfo{person}{Arthur Mensch}, \bibinfo{person}{Katherine Millican}, \bibinfo{person}{Malcolm Reynolds}, \bibinfo{person}{Roman Ring}, \bibinfo{person}{Eliza Rutherford}, \bibinfo{person}{Serkan Cabi}, \bibinfo{person}{Tengda Han}, \bibinfo{person}{Zhitao Gong}, \bibinfo{person}{Sina Samangooei}, \bibinfo{person}{Marianne Monteiro}, \bibinfo{person}{Jacob~L. Menick}, \bibinfo{person}{Sebastian Borgeaud}, \bibinfo{person}{Andy Brock}, \bibinfo{person}{Aida Nematzadeh}, \bibinfo{person}{Sahand Sharifzadeh}, \bibinfo{person}{Mikołaj Bińkowski}, \bibinfo{person}{Ricardo Barreira}, \bibinfo{person}{Oriol Vinyals}, \bibinfo{person}{Andrew Zisserman}, {and} \bibinfo{person}{Karén Simonyan}.} \bibinfo{year}{2022}\natexlab{}.
\newblock \showarticletitle{Flamingo: a {Visual} {Language} {Model} for {Few}-{Shot} {Learning}}.
\newblock \bibinfo{journal}{\emph{Advances in Neural Information Processing Systems}}  \bibinfo{volume}{35} (\bibinfo{date}{Dec.} \bibinfo{year}{2022}), \bibinfo{pages}{23716--23736}.
\newblock
\urldef\tempurl%
\url{https://proceedings.neurips.cc/paper_files/paper/2022/hash/960a172bc7fbf0177ccccbb411a7d800-Abstract-Conference.html}
\showURL{%
\tempurl}


\bibitem[Allal et~al\mbox{.}(2025)]%
        {allal_smollm2_2025}
\bibfield{author}{\bibinfo{person}{Loubna~Ben Allal}, \bibinfo{person}{Anton Lozhkov}, \bibinfo{person}{Elie Bakouch}, \bibinfo{person}{Gabriel~Martín Blázquez}, \bibinfo{person}{Guilherme Penedo}, \bibinfo{person}{Lewis Tunstall}, \bibinfo{person}{Andrés Marafioti}, \bibinfo{person}{Hynek Kydlíček}, \bibinfo{person}{Agustín~Piqueres Lajarín}, \bibinfo{person}{Vaibhav Srivastav}, \bibinfo{person}{Joshua Lochner}, \bibinfo{person}{Caleb Fahlgren}, \bibinfo{person}{Xuan-Son Nguyen}, \bibinfo{person}{Clémentine Fourrier}, \bibinfo{person}{Ben Burtenshaw}, \bibinfo{person}{Hugo Larcher}, \bibinfo{person}{Haojun Zhao}, \bibinfo{person}{Cyril Zakka}, \bibinfo{person}{Mathieu Morlon}, \bibinfo{person}{Colin Raffel}, \bibinfo{person}{Leandro~von Werra}, {and} \bibinfo{person}{Thomas Wolf}.} \bibinfo{year}{2025}\natexlab{}.
\newblock \bibinfo{title}{{SmolLM2}: {When} {Smol} {Goes} {Big} -- {Data}-{Centric} {Training} of a {Small} {Language} {Model}}.
\newblock
\href{https://doi.org/10.48550/arXiv.2502.02737}{doi:\nolinkurl{10.48550/arXiv.2502.02737}}
\newblock
\shownote{arXiv:2502.02737 [cs] version: 1}.


\bibitem[Apostolidis et~al\mbox{.}(2021)]%
        {apostolidis_video_2021}
\bibfield{author}{\bibinfo{person}{Evlampios Apostolidis}, \bibinfo{person}{Eleni Adamantidou}, \bibinfo{person}{Alexandros~I. Metsai}, \bibinfo{person}{Vasileios Mezaris}, {and} \bibinfo{person}{Ioannis Patras}.} \bibinfo{year}{2021}\natexlab{}.
\newblock \showarticletitle{Video {Summarization} {Using} {Deep} {Neural} {Networks}: {A} {Survey}}.
\newblock \bibinfo{journal}{\emph{Proc. IEEE}} \bibinfo{volume}{109}, \bibinfo{number}{11} (\bibinfo{date}{Nov.} \bibinfo{year}{2021}), \bibinfo{pages}{1838--1863}.
\newblock
\showISSN{1558-2256}
\href{https://doi.org/10.1109/JPROC.2021.3117472}{doi:\nolinkurl{10.1109/JPROC.2021.3117472}}
\newblock
\shownote{Conference Name: Proceedings of the IEEE}.


\bibitem[Beltagy et~al\mbox{.}(2020)]%
        {beltagy_longformer_2020}
\bibfield{author}{\bibinfo{person}{Iz Beltagy}, \bibinfo{person}{Matthew~E. Peters}, {and} \bibinfo{person}{Arman Cohan}.} \bibinfo{year}{2020}\natexlab{}.
\newblock \bibinfo{title}{Longformer: {The} {Long}-{Document} {Transformer}}.
\newblock
\href{https://doi.org/10.48550/arXiv.2004.05150}{doi:\nolinkurl{10.48550/arXiv.2004.05150}}
\newblock
\shownote{arXiv:2004.05150 [cs]}.


\bibitem[Castro et~al\mbox{.}(2019)]%
        {castro_towards_2019}
\bibfield{author}{\bibinfo{person}{Santiago Castro}, \bibinfo{person}{Devamanyu Hazarika}, \bibinfo{person}{Verónica Pérez-Rosas}, \bibinfo{person}{Roger Zimmermann}, \bibinfo{person}{Rada Mihalcea}, {and} \bibinfo{person}{Soujanya Poria}.} \bibinfo{year}{2019}\natexlab{}.
\newblock \showarticletitle{Towards {Multimodal} {Sarcasm} {Detection} ({An} \_Obviously\_ {Perfect} {Paper})}. In \bibinfo{booktitle}{\emph{Proceedings of the 57th {Annual} {Meeting} of the {Association} for {Computational} {Linguistics}}}, \bibfield{editor}{\bibinfo{person}{Anna Korhonen}, \bibinfo{person}{David Traum}, {and} \bibinfo{person}{Lluís Màrquez}} (Eds.). \bibinfo{publisher}{Association for Computational Linguistics}, \bibinfo{address}{Florence, Italy}, \bibinfo{pages}{4619--4629}.
\newblock
\href{https://doi.org/10.18653/v1/P19-1455}{doi:\nolinkurl{10.18653/v1/P19-1455}}


\bibitem[Chen et~al\mbox{.}(2023)]%
        {chen_can_2023}
\bibfield{author}{\bibinfo{person}{Yang Chen}, \bibinfo{person}{Hexiang Hu}, \bibinfo{person}{Yi Luan}, \bibinfo{person}{Haitian Sun}, \bibinfo{person}{Soravit Changpinyo}, \bibinfo{person}{Alan Ritter}, {and} \bibinfo{person}{Ming-Wei Chang}.} \bibinfo{year}{2023}\natexlab{}.
\newblock \showarticletitle{Can {Pre}-trained {Vision} and {Language} {Models} {Answer} {Visual} {Information}-{Seeking} {Questions}?}
\newblock
\urldef\tempurl%
\url{https://openreview.net/forum?id=3MEV3aIDDq}
\showURL{%
\tempurl}


\bibitem[Dao(2023)]%
        {dao_flashattention-2_2023}
\bibfield{author}{\bibinfo{person}{Tri Dao}.} \bibinfo{year}{2023}\natexlab{}.
\newblock \bibinfo{title}{{FlashAttention}-2: {Faster} {Attention} with {Better} {Parallelism} and {Work} {Partitioning}}.
\newblock
\href{https://doi.org/10.48550/arXiv.2307.08691}{doi:\nolinkurl{10.48550/arXiv.2307.08691}}
\newblock
\shownote{arXiv:2307.08691 [cs]}.


\bibitem[Dao et~al\mbox{.}(2022)]%
        {dao_flashattention_2022}
\bibfield{author}{\bibinfo{person}{Tri Dao}, \bibinfo{person}{Dan Fu}, \bibinfo{person}{Stefano Ermon}, \bibinfo{person}{Atri Rudra}, {and} \bibinfo{person}{Christopher Ré}.} \bibinfo{year}{2022}\natexlab{}.
\newblock \showarticletitle{{FlashAttention}: {Fast} and {Memory}-{Efficient} {Exact} {Attention} with {IO}-{Awareness}}.
\newblock \bibinfo{journal}{\emph{Advances in Neural Information Processing Systems}}  \bibinfo{volume}{35} (\bibinfo{date}{Dec.} \bibinfo{year}{2022}), \bibinfo{pages}{16344--16359}.
\newblock
\urldef\tempurl%
\url{https://proceedings.neurips.cc/paper_files/paper/2022/hash/67d57c32e20fd0a7a302cb81d36e40d5-Abstract-Conference.html}
\showURL{%
\tempurl}


\bibitem[Darcet et~al\mbox{.}(2023)]%
        {darcet_vision_2023}
\bibfield{author}{\bibinfo{person}{Timothée Darcet}, \bibinfo{person}{Maxime Oquab}, \bibinfo{person}{Julien Mairal}, {and} \bibinfo{person}{Piotr Bojanowski}.} \bibinfo{year}{2023}\natexlab{}.
\newblock \showarticletitle{Vision {Transformers} {Need} {Registers}}.
\newblock
\urldef\tempurl%
\url{https://openreview.net/forum?id=2dnO3LLiJ1}
\showURL{%
\tempurl}


\bibitem[Deng et~al\mbox{.}(2025)]%
        {deng_words_2025}
\bibfield{author}{\bibinfo{person}{Ailin Deng}, \bibinfo{person}{Tri Cao}, \bibinfo{person}{Zhirui Chen}, {and} \bibinfo{person}{Bryan Hooi}.} \bibinfo{year}{2025}\natexlab{}.
\newblock \bibinfo{title}{Words or {Vision}: {Do} {Vision}-{Language} {Models} {Have} {Blind} {Faith} in {Text}?}
\newblock
\href{https://doi.org/10.48550/arXiv.2503.02199}{doi:\nolinkurl{10.48550/arXiv.2503.02199}}
\newblock
\shownote{arXiv:2503.02199 [cs]}.


\bibitem[Gigant et~al\mbox{.}(2023)]%
        {gigant_tib_2023}
\bibfield{author}{\bibinfo{person}{Théo Gigant}, \bibinfo{person}{Frederic Dufaux}, \bibinfo{person}{Camille Guinaudeau}, {and} \bibinfo{person}{Marc Décombas}.} \bibinfo{year}{2023}\natexlab{}.
\newblock \showarticletitle{{TIB}: {A} {Dataset} for {Abstractive} {Summarization} of {Long} {Multimodal} {Videoconference} {Records}}. In \bibinfo{booktitle}{\emph{Proceedings of the 20th {International} {Conference} on {Content}-based {Multimedia} {Indexing}}} \emph{(\bibinfo{series}{{CBMI} '23})}. \bibinfo{publisher}{Association for Computing Machinery}, \bibinfo{address}{New York, NY, USA}, \bibinfo{pages}{61--70}.
\newblock
\showISBNx{9798400709128}
\href{https://doi.org/10.1145/3617233.3617238}{doi:\nolinkurl{10.1145/3617233.3617238}}


\bibitem[Gigant et~al\mbox{.}(2024)]%
        {gigant_mitigating_2024}
\bibfield{author}{\bibinfo{person}{Théo Gigant}, \bibinfo{person}{Camille Guinaudeau}, \bibinfo{person}{Marc Decombas}, {and} \bibinfo{person}{Frederic Dufaux}.} \bibinfo{year}{2024}\natexlab{}.
\newblock \showarticletitle{Mitigating the {Impact} of {Reference} {Quality} on {Evaluation} of {Summarization} {Systems} with {Reference}-{Free} {Metrics}}. In \bibinfo{booktitle}{\emph{Proceedings of the 2024 {Conference} on {Empirical} {Methods} in {Natural} {Language} {Processing}}}, \bibfield{editor}{\bibinfo{person}{Yaser Al-Onaizan}, \bibinfo{person}{Mohit Bansal}, {and} \bibinfo{person}{Yun-Nung Chen}} (Eds.). \bibinfo{publisher}{Association for Computational Linguistics}, \bibinfo{address}{Miami, Florida, USA}, \bibinfo{pages}{19355--19368}.
\newblock
\href{https://doi.org/10.18653/v1/2024.emnlp-main.1078}{doi:\nolinkurl{10.18653/v1/2024.emnlp-main.1078}}


\bibitem[Goh et~al\mbox{.}(2021)]%
        {goh_multimodal_2021}
\bibfield{author}{\bibinfo{person}{Gabriel Goh}, \bibinfo{person}{Nick~Cammarata †}, \bibinfo{person}{Chelsea~Voss †}, \bibinfo{person}{Shan Carter}, \bibinfo{person}{Michael Petrov}, \bibinfo{person}{Ludwig Schubert}, \bibinfo{person}{Alec Radford}, {and} \bibinfo{person}{Chris Olah}.} \bibinfo{year}{2021}\natexlab{}.
\newblock \showarticletitle{Multimodal {Neurons} in {Artificial} {Neural} {Networks}}.
\newblock \bibinfo{journal}{\emph{Distill}} \bibinfo{volume}{6}, \bibinfo{number}{3} (\bibinfo{date}{March} \bibinfo{year}{2021}), \bibinfo{pages}{e30}.
\newblock
\showISSN{2476-0757}
\href{https://doi.org/10.23915/distill.00030}{doi:\nolinkurl{10.23915/distill.00030}}


\bibitem[Guo et~al\mbox{.}(2022)]%
        {guo_longt5_2022}
\bibfield{author}{\bibinfo{person}{Mandy Guo}, \bibinfo{person}{Joshua Ainslie}, \bibinfo{person}{David Uthus}, \bibinfo{person}{Santiago Ontanon}, \bibinfo{person}{Jianmo Ni}, \bibinfo{person}{Yun-Hsuan Sung}, {and} \bibinfo{person}{Yinfei Yang}.} \bibinfo{year}{2022}\natexlab{}.
\newblock \showarticletitle{{LongT5}: {Efficient} {Text}-{To}-{Text} {Transformer} for {Long} {Sequences}}. In \bibinfo{booktitle}{\emph{Findings of the {Association} for {Computational} {Linguistics}: {NAACL} 2022}}, \bibfield{editor}{\bibinfo{person}{Marine Carpuat}, \bibinfo{person}{Marie-Catherine de~Marneffe}, {and} \bibinfo{person}{Ivan~Vladimir Meza~Ruiz}} (Eds.). \bibinfo{publisher}{Association for Computational Linguistics}, \bibinfo{address}{Seattle, United States}, \bibinfo{pages}{724--736}.
\newblock
\href{https://doi.org/10.18653/v1/2022.findings-naacl.55}{doi:\nolinkurl{10.18653/v1/2022.findings-naacl.55}}


\bibitem[Kim et~al\mbox{.}(2022)]%
        {kim_ocr-free_2022}
\bibfield{author}{\bibinfo{person}{Geewook Kim}, \bibinfo{person}{Teakgyu Hong}, \bibinfo{person}{Moonbin Yim}, \bibinfo{person}{JeongYeon Nam}, \bibinfo{person}{Jinyoung Park}, \bibinfo{person}{Jinyeong Yim}, \bibinfo{person}{Wonseok Hwang}, \bibinfo{person}{Sangdoo Yun}, \bibinfo{person}{Dongyoon Han}, {and} \bibinfo{person}{Seunghyun Park}.} \bibinfo{year}{2022}\natexlab{}.
\newblock \showarticletitle{{OCR}-{Free} {Document} {Understanding} {Transformer}}. In \bibinfo{booktitle}{\emph{Computer {Vision} – {ECCV} 2022}}, \bibfield{editor}{\bibinfo{person}{Shai Avidan}, \bibinfo{person}{Gabriel Brostow}, \bibinfo{person}{Moustapha Cissé}, \bibinfo{person}{Giovanni~Maria Farinella}, {and} \bibinfo{person}{Tal Hassner}} (Eds.). \bibinfo{publisher}{Springer Nature Switzerland}, \bibinfo{address}{Cham}, \bibinfo{pages}{498--517}.
\newblock
\showISBNx{978-3-031-19815-1}
\href{https://doi.org/10.1007/978-3-031-19815-1_29}{doi:\nolinkurl{10.1007/978-3-031-19815-1_29}}


\bibitem[Koh et~al\mbox{.}(2022)]%
        {koh_how_2022}
\bibfield{author}{\bibinfo{person}{Huan~Yee Koh}, \bibinfo{person}{Jiaxin Ju}, \bibinfo{person}{He Zhang}, \bibinfo{person}{Ming Liu}, {and} \bibinfo{person}{Shirui Pan}.} \bibinfo{year}{2022}\natexlab{}.
\newblock \showarticletitle{How {Far} are {We} from {Robust} {Long} {Abstractive} {Summarization}?}. In \bibinfo{booktitle}{\emph{Proceedings of the 2022 {Conference} on {Empirical} {Methods} in {Natural} {Language} {Processing}}}, \bibfield{editor}{\bibinfo{person}{Yoav Goldberg}, \bibinfo{person}{Zornitsa Kozareva}, {and} \bibinfo{person}{Yue Zhang}} (Eds.). \bibinfo{publisher}{Association for Computational Linguistics}, \bibinfo{address}{Abu Dhabi, United Arab Emirates}, \bibinfo{pages}{2682--2698}.
\newblock
\href{https://doi.org/10.18653/v1/2022.emnlp-main.172}{doi:\nolinkurl{10.18653/v1/2022.emnlp-main.172}}


\bibitem[Laurençon et~al\mbox{.}(2024)]%
        {laurencon_building_2024}
\bibfield{author}{\bibinfo{person}{Hugo Laurençon}, \bibinfo{person}{Andrés Marafioti}, \bibinfo{person}{Victor Sanh}, {and} \bibinfo{person}{Leo Tronchon}.} \bibinfo{year}{2024}\natexlab{}.
\newblock \showarticletitle{Building and better understanding vision-language models: insights and future directions}.
\newblock
\urldef\tempurl%
\url{https://openreview.net/forum?id=iSL0FHZStr}
\showURL{%
\tempurl}


\bibitem[Laurençon et~al\mbox{.}(2025)]%
        {laurencon_what_2025}
\bibfield{author}{\bibinfo{person}{Hugo Laurençon}, \bibinfo{person}{Leo Tronchon}, \bibinfo{person}{Matthieu Cord}, {and} \bibinfo{person}{Victor Sanh}.} \bibinfo{year}{2025}\natexlab{}.
\newblock \showarticletitle{What matters when building vision-language models?}
\newblock \bibinfo{journal}{\emph{Advances in Neural Information Processing Systems}}  \bibinfo{volume}{37} (\bibinfo{date}{Jan.} \bibinfo{year}{2025}), \bibinfo{pages}{87874--87907}.
\newblock
\urldef\tempurl%
\url{https://proceedings.neurips.cc/paper_files/paper/2024/hash/a03037317560b8c5f2fb4b6466d4c439-Abstract-Conference.html}
\showURL{%
\tempurl}


\bibitem[Lerner et~al\mbox{.}(2022)]%
        {lerner_viquae_2022}
\bibfield{author}{\bibinfo{person}{Paul Lerner}, \bibinfo{person}{Olivier Ferret}, \bibinfo{person}{Camille Guinaudeau}, \bibinfo{person}{Hervé Le~Borgne}, \bibinfo{person}{Romaric Besançon}, \bibinfo{person}{Jose~G. Moreno}, {and} \bibinfo{person}{Jesús Lovón~Melgarejo}.} \bibinfo{year}{2022}\natexlab{}.
\newblock \showarticletitle{{ViQuAE}, a {Dataset} for {Knowledge}-based {Visual} {Question} {Answering} about {Named} {Entities}}. In \bibinfo{booktitle}{\emph{Proceedings of the 45th {International} {ACM} {SIGIR} {Conference} on {Research} and {Development} in {Information} {Retrieval}}} \emph{(\bibinfo{series}{{SIGIR} '22})}. \bibinfo{publisher}{Association for Computing Machinery}, \bibinfo{address}{New York, NY, USA}, \bibinfo{pages}{3108--3120}.
\newblock
\showISBNx{978-1-4503-8732-3}
\href{https://doi.org/10.1145/3477495.3531753}{doi:\nolinkurl{10.1145/3477495.3531753}}


\bibitem[Lewis et~al\mbox{.}(2020)]%
        {lewis_bart_2020}
\bibfield{author}{\bibinfo{person}{Mike Lewis}, \bibinfo{person}{Yinhan Liu}, \bibinfo{person}{Naman Goyal}, \bibinfo{person}{Marjan Ghazvininejad}, \bibinfo{person}{Abdelrahman Mohamed}, \bibinfo{person}{Omer Levy}, \bibinfo{person}{Veselin Stoyanov}, {and} \bibinfo{person}{Luke Zettlemoyer}.} \bibinfo{year}{2020}\natexlab{}.
\newblock \showarticletitle{{BART}: {Denoising} {Sequence}-to-{Sequence} {Pre}-training for {Natural} {Language} {Generation}, {Translation}, and {Comprehension}}. In \bibinfo{booktitle}{\emph{Proceedings of the 58th {Annual} {Meeting} of the {Association} for {Computational} {Linguistics}}}, \bibfield{editor}{\bibinfo{person}{Dan Jurafsky}, \bibinfo{person}{Joyce Chai}, \bibinfo{person}{Natalie Schluter}, {and} \bibinfo{person}{Joel Tetreault}} (Eds.). \bibinfo{publisher}{Association for Computational Linguistics}, \bibinfo{address}{Online}, \bibinfo{pages}{7871--7880}.
\newblock
\href{https://doi.org/10.18653/v1/2020.acl-main.703}{doi:\nolinkurl{10.18653/v1/2020.acl-main.703}}


\bibitem[Li et~al\mbox{.}(2024)]%
        {li_long-context_2024}
\bibfield{author}{\bibinfo{person}{Tianle Li}, \bibinfo{person}{Ge Zhang}, \bibinfo{person}{Quy~Duc Do}, \bibinfo{person}{Xiang Yue}, {and} \bibinfo{person}{Wenhu Chen}.} \bibinfo{year}{2024}\natexlab{}.
\newblock \showarticletitle{Long-context {LLMs} {Struggle} with {Long} {In}-context {Learning}}.
\newblock \bibinfo{journal}{\emph{CoRR}} (\bibinfo{date}{Jan.} \bibinfo{year}{2024}).
\newblock
\urldef\tempurl%
\url{https://openreview.net/forum?id=CFYzMJJB6a}
\showURL{%
\tempurl}


\bibitem[Liang et~al\mbox{.}(2022)]%
        {liang_mind_2022}
\bibfield{author}{\bibinfo{person}{Victor~Weixin Liang}, \bibinfo{person}{Yuhui Zhang}, \bibinfo{person}{Yongchan Kwon}, \bibinfo{person}{Serena Yeung}, {and} \bibinfo{person}{James~Y. Zou}.} \bibinfo{year}{2022}\natexlab{}.
\newblock \showarticletitle{Mind the {Gap}: {Understanding} the {Modality} {Gap} in {Multi}-modal {Contrastive} {Representation} {Learning}}.
\newblock \bibinfo{journal}{\emph{Advances in Neural Information Processing Systems}}  \bibinfo{volume}{35} (\bibinfo{date}{Dec.} \bibinfo{year}{2022}), \bibinfo{pages}{17612--17625}.
\newblock
\urldef\tempurl%
\url{https://proceedings.neurips.cc/paper_files/paper/2022/hash/702f4db7543a7432431df588d57bc7c9-Abstract-Conference.html}
\showURL{%
\tempurl}


\bibitem[Lin et~al\mbox{.}(2024)]%
        {lin_video-llava_2024}
\bibfield{author}{\bibinfo{person}{Bin Lin}, \bibinfo{person}{Yang Ye}, \bibinfo{person}{Bin Zhu}, \bibinfo{person}{Jiaxi Cui}, \bibinfo{person}{Munan Ning}, \bibinfo{person}{Peng Jin}, {and} \bibinfo{person}{Li Yuan}.} \bibinfo{year}{2024}\natexlab{}.
\newblock \showarticletitle{Video-{LLaVA}: {Learning} {United} {Visual} {Representation} by {Alignment} {Before} {Projection}}. In \bibinfo{booktitle}{\emph{Proceedings of the 2024 {Conference} on {Empirical} {Methods} in {Natural} {Language} {Processing}}}, \bibfield{editor}{\bibinfo{person}{Yaser Al-Onaizan}, \bibinfo{person}{Mohit Bansal}, {and} \bibinfo{person}{Yun-Nung Chen}} (Eds.). \bibinfo{publisher}{Association for Computational Linguistics}, \bibinfo{address}{Miami, Florida, USA}, \bibinfo{pages}{5971--5984}.
\newblock
\href{https://doi.org/10.18653/v1/2024.emnlp-main.342}{doi:\nolinkurl{10.18653/v1/2024.emnlp-main.342}}


\bibitem[Lin(2004)]%
        {lin_rouge_2004}
\bibfield{author}{\bibinfo{person}{Chin-Yew Lin}.} \bibinfo{year}{2004}\natexlab{}.
\newblock \showarticletitle{{ROUGE}: {A} {Package} for {Automatic} {Evaluation} of {Summaries}}. In \bibinfo{booktitle}{\emph{Text {Summarization} {Branches} {Out}}}. \bibinfo{publisher}{Association for Computational Linguistics}, \bibinfo{address}{Barcelona, Spain}, \bibinfo{pages}{74--81}.
\newblock
\urldef\tempurl%
\url{https://aclanthology.org/W04-1013}
\showURL{%
\tempurl}


\bibitem[Liu et~al\mbox{.}(2025)]%
        {liu_what_2025}
\bibfield{author}{\bibinfo{person}{Dongqi Liu}, \bibinfo{person}{Chenxi Whitehouse}, \bibinfo{person}{Xi Yu}, \bibinfo{person}{Louis Mahon}, \bibinfo{person}{Rohit Saxena}, \bibinfo{person}{Zheng Zhao}, \bibinfo{person}{Yifu Qiu}, \bibinfo{person}{Mirella Lapata}, {and} \bibinfo{person}{Vera Demberg}.} \bibinfo{year}{2025}\natexlab{}.
\newblock \bibinfo{title}{What {Is} {That} {Talk} {About}? {A} {Video}-to-{Text} {Summarization} {Dataset} for {Scientific} {Presentations}}.
\newblock
\href{https://doi.org/10.48550/arXiv.2502.08279}{doi:\nolinkurl{10.48550/arXiv.2502.08279}}
\newblock
\shownote{arXiv:2502.08279 [cs]}.


\bibitem[Liu et~al\mbox{.}(2023)]%
        {liu_visual_2023}
\bibfield{author}{\bibinfo{person}{Haotian Liu}, \bibinfo{person}{Chunyuan Li}, \bibinfo{person}{Qingyang Wu}, {and} \bibinfo{person}{Yong~Jae Lee}.} \bibinfo{year}{2023}\natexlab{}.
\newblock \showarticletitle{Visual {Instruction} {Tuning}}.
\newblock \bibinfo{journal}{\emph{Advances in Neural Information Processing Systems}}  \bibinfo{volume}{36} (\bibinfo{date}{Dec.} \bibinfo{year}{2023}), \bibinfo{pages}{34892--34916}.
\newblock
\urldef\tempurl%
\url{https://papers.nips.cc/paper_files/paper/2023/hash/6dcf277ea32ce3288914faf369fe6de0-Abstract-Conference.html}
\showURL{%
\tempurl}


\bibitem[Liu et~al\mbox{.}(2018)]%
        {liu_learn_2018}
\bibfield{author}{\bibinfo{person}{Kuan Liu}, \bibinfo{person}{Yanen Li}, \bibinfo{person}{Ning Xu}, {and} \bibinfo{person}{Prem Natarajan}.} \bibinfo{year}{2018}\natexlab{}.
\newblock \bibinfo{title}{Learn to {Combine} {Modalities} in {Multimodal} {Deep} {Learning}}.
\newblock
\href{https://doi.org/10.48550/arXiv.1805.11730}{doi:\nolinkurl{10.48550/arXiv.1805.11730}}
\newblock
\shownote{arXiv:1805.11730 [stat]}.


\bibitem[Liu et~al\mbox{.}(2024)]%
        {liu_insight_2024}
\bibfield{author}{\bibinfo{person}{Xiaoyuan Liu}, \bibinfo{person}{Wenxuan Wang}, \bibinfo{person}{Youliang Yuan}, \bibinfo{person}{Jen-tse Huang}, \bibinfo{person}{Qiuzhi Liu}, \bibinfo{person}{Pinjia He}, {and} \bibinfo{person}{Zhaopeng Tu}.} \bibinfo{year}{2024}\natexlab{}.
\newblock \bibinfo{title}{Insight {Over} {Sight}? {Exploring} the {Vision}-{Knowledge} {Conflicts} in {Multimodal} {LLMs}}.
\newblock
\href{https://doi.org/10.48550/arXiv.2410.08145}{doi:\nolinkurl{10.48550/arXiv.2410.08145}}
\newblock
\shownote{arXiv:2410.08145 [cs]}.


\bibitem[Luo et~al\mbox{.}(2025)]%
        {luo_valley_2025}
\bibfield{author}{\bibinfo{person}{Ruipu Luo}, \bibinfo{person}{Ziwang Zhao}, \bibinfo{person}{Min Yang}, \bibinfo{person}{Zheming Yang}, \bibinfo{person}{Minghui Qiu}, \bibinfo{person}{Tao Wang}, \bibinfo{person}{Zhongyu Wei}, \bibinfo{person}{Yanhao Wang}, {and} \bibinfo{person}{Cen Chen}.} \bibinfo{year}{2025}\natexlab{}.
\newblock \bibinfo{title}{Valley: {Video} {Assistant} with {Large} {Language} model {Enhanced} {abilitY}}.
\newblock
\href{https://doi.org/10.48550/arXiv.2306.07207}{doi:\nolinkurl{10.48550/arXiv.2306.07207}}
\newblock
\shownote{arXiv:2306.07207 [cs]}.


\bibitem[Materzyńska et~al\mbox{.}(2022)]%
        {materzynska_disentangling_2022}
\bibfield{author}{\bibinfo{person}{Joanna Materzyńska}, \bibinfo{person}{Antonio Torralba}, {and} \bibinfo{person}{David Bau}.} \bibinfo{year}{2022}\natexlab{}.
\newblock \showarticletitle{Disentangling {Visual} and {Written} {Concepts} in {CLIP}}. \bibinfo{pages}{16410--16419}.
\newblock
\urldef\tempurl%
\url{https://openaccess.thecvf.com/content/CVPR2022/html/Materzynska_Disentangling_Visual_and_Written_Concepts_in_CLIP_CVPR_2022_paper.html}
\showURL{%
\tempurl}


\bibitem[McKinzie et~al\mbox{.}(2025)]%
        {mckinzie_mm1_2025}
\bibfield{author}{\bibinfo{person}{Brandon McKinzie}, \bibinfo{person}{Zhe Gan}, \bibinfo{person}{Jean-Philippe Fauconnier}, \bibinfo{person}{Sam Dodge}, \bibinfo{person}{Bowen Zhang}, \bibinfo{person}{Philipp Dufter}, \bibinfo{person}{Dhruti Shah}, \bibinfo{person}{Xianzhi Du}, \bibinfo{person}{Futang Peng}, \bibinfo{person}{Anton Belyi}, \bibinfo{person}{Haotian Zhang}, \bibinfo{person}{Karanjeet Singh}, \bibinfo{person}{Doug Kang}, \bibinfo{person}{Hongyu Hè}, \bibinfo{person}{Max Schwarzer}, \bibinfo{person}{Tom Gunter}, \bibinfo{person}{Xiang Kong}, \bibinfo{person}{Aonan Zhang}, \bibinfo{person}{Jianyu Wang}, \bibinfo{person}{Chong Wang}, \bibinfo{person}{Nan Du}, \bibinfo{person}{Tao Lei}, \bibinfo{person}{Sam Wiseman}, \bibinfo{person}{Mark Lee}, \bibinfo{person}{Zirui Wang}, \bibinfo{person}{Ruoming Pang}, \bibinfo{person}{Peter Grasch}, \bibinfo{person}{Alexander Toshev}, {and} \bibinfo{person}{Yinfei Yang}.} \bibinfo{year}{2025}\natexlab{}.
\newblock \showarticletitle{{MM1}: {Methods}, {Analysis} and {Insights} from {Multimodal} {LLM} {Pre}-training}. In \bibinfo{booktitle}{\emph{Computer {Vision} – {ECCV} 2024}}, \bibfield{editor}{\bibinfo{person}{Aleš Leonardis}, \bibinfo{person}{Elisa Ricci}, \bibinfo{person}{Stefan Roth}, \bibinfo{person}{Olga Russakovsky}, \bibinfo{person}{Torsten Sattler}, {and} \bibinfo{person}{Gül Varol}} (Eds.). \bibinfo{publisher}{Springer Nature Switzerland}, \bibinfo{address}{Cham}, \bibinfo{pages}{304--323}.
\newblock
\showISBNx{978-3-031-73397-0}
\href{https://doi.org/10.1007/978-3-031-73397-0_18}{doi:\nolinkurl{10.1007/978-3-031-73397-0_18}}


\bibitem[Peng et~al\mbox{.}(2023)]%
        {peng_yarn_2023}
\bibfield{author}{\bibinfo{person}{Bowen Peng}, \bibinfo{person}{Jeffrey Quesnelle}, \bibinfo{person}{Honglu Fan}, {and} \bibinfo{person}{Enrico Shippole}.} \bibinfo{year}{2023}\natexlab{}.
\newblock \showarticletitle{{YaRN}: {Efficient} {Context} {Window} {Extension} of {Large} {Language} {Models}}.
\newblock
\urldef\tempurl%
\url{https://openreview.net/forum?id=wHBfxhZu1u}
\showURL{%
\tempurl}


\bibitem[Radford et~al\mbox{.}(2021)]%
        {radford_learning_2021}
\bibfield{author}{\bibinfo{person}{Alec Radford}, \bibinfo{person}{Jong~Wook Kim}, \bibinfo{person}{Chris Hallacy}, \bibinfo{person}{Aditya Ramesh}, \bibinfo{person}{Gabriel Goh}, \bibinfo{person}{Sandhini Agarwal}, \bibinfo{person}{Girish Sastry}, \bibinfo{person}{Amanda Askell}, \bibinfo{person}{Pamela Mishkin}, \bibinfo{person}{Jack Clark}, \bibinfo{person}{Gretchen Krueger}, {and} \bibinfo{person}{Ilya Sutskever}.} \bibinfo{year}{2021}\natexlab{}.
\newblock \showarticletitle{Learning {Transferable} {Visual} {Models} {From} {Natural} {Language} {Supervision}}. In \bibinfo{booktitle}{\emph{Proceedings of the 38th {International} {Conference} on {Machine} {Learning}}}. \bibinfo{publisher}{PMLR}, \bibinfo{pages}{8748--8763}.
\newblock
\urldef\tempurl%
\url{https://proceedings.mlr.press/v139/radford21a.html}
\showURL{%
\tempurl}
\newblock
\shownote{ISSN: 2640-3498}.


\bibitem[Raffel et~al\mbox{.}(2020)]%
        {raffel_exploring_2020}
\bibfield{author}{\bibinfo{person}{Colin Raffel}, \bibinfo{person}{Noam Shazeer}, \bibinfo{person}{Adam Roberts}, \bibinfo{person}{Katherine Lee}, \bibinfo{person}{Sharan Narang}, \bibinfo{person}{Michael Matena}, \bibinfo{person}{Yanqi Zhou}, \bibinfo{person}{Wei Li}, {and} \bibinfo{person}{Peter~J. Liu}.} \bibinfo{year}{2020}\natexlab{}.
\newblock \showarticletitle{Exploring the {Limits} of {Transfer} {Learning} with a {Unified} {Text}-to-{Text} {Transformer}}.
\newblock \bibinfo{journal}{\emph{Journal of Machine Learning Research}} \bibinfo{volume}{21}, \bibinfo{number}{140} (\bibinfo{year}{2020}), \bibinfo{pages}{1--67}.
\newblock
\showISSN{1533-7928}
\urldef\tempurl%
\url{http://jmlr.org/papers/v21/20-074.html}
\showURL{%
\tempurl}


\bibitem[Templeton et~al\mbox{.}(2024)]%
        {templeton_scaling_2024}
\bibfield{author}{\bibinfo{person}{Adly Templeton}, \bibinfo{person}{Tom Conerly}, \bibinfo{person}{Jonathan Marcus}, \bibinfo{person}{Jack Lindsey}, \bibinfo{person}{Trenton Bricken}, \bibinfo{person}{Brian Chen}, \bibinfo{person}{Adam Pearce}, \bibinfo{person}{Craig Citro}, \bibinfo{person}{Emmanuel Ameisen}, \bibinfo{person}{Andy Jones}, {and} \bibinfo{person}{{others}}.} \bibinfo{year}{2024}\natexlab{}.
\newblock \bibinfo{title}{Scaling monosemanticity: extracting interpretable features from {Claude} 3 sonnet.}
\newblock


\bibitem[Vasilyev and Bohannon(2021)]%
        {vasilyev_estime_2021}
\bibfield{author}{\bibinfo{person}{Oleg Vasilyev} {and} \bibinfo{person}{John Bohannon}.} \bibinfo{year}{2021}\natexlab{}.
\newblock \showarticletitle{{ESTIME}: {Estimation} of {Summary}-to-{Text} {Inconsistency} by {Mismatched} {Embeddings}}. In \bibinfo{booktitle}{\emph{Proceedings of the 2nd {Workshop} on {Evaluation} and {Comparison} of {NLP} {Systems}}}, \bibfield{editor}{\bibinfo{person}{Yang Gao}, \bibinfo{person}{Steffen Eger}, \bibinfo{person}{Wei Zhao}, \bibinfo{person}{Piyawat Lertvittayakumjorn}, {and} \bibinfo{person}{Marina Fomicheva}} (Eds.). \bibinfo{publisher}{Association for Computational Linguistics}, \bibinfo{address}{Punta Cana, Dominican Republic}, \bibinfo{pages}{94--103}.
\newblock
\href{https://doi.org/10.18653/v1/2021.eval4nlp-1.10}{doi:\nolinkurl{10.18653/v1/2021.eval4nlp-1.10}}


\bibitem[Vaswani et~al\mbox{.}(2017)]%
        {vaswani_attention_2017}
\bibfield{author}{\bibinfo{person}{Ashish Vaswani}, \bibinfo{person}{Noam Shazeer}, \bibinfo{person}{Niki Parmar}, \bibinfo{person}{Jakob Uszkoreit}, \bibinfo{person}{Llion Jones}, \bibinfo{person}{Aidan~N Gomez}, \bibinfo{person}{Ł~ukasz Kaiser}, {and} \bibinfo{person}{Illia Polosukhin}.} \bibinfo{year}{2017}\natexlab{}.
\newblock \showarticletitle{Attention is {All} you {Need}}. In \bibinfo{booktitle}{\emph{Advances in {Neural} {Information} {Processing} {Systems}}}, Vol.~\bibinfo{volume}{30}. \bibinfo{publisher}{Curran Associates, Inc.}
\newblock
\urldef\tempurl%
\url{https://proceedings.neurips.cc/paper/2017/hash/3f5ee243547dee91fbd053c1c4a845aa-Abstract.html}
\showURL{%
\tempurl}


\bibitem[Wang et~al\mbox{.}(2024)]%
        {wang_qwen2-vl_2024}
\bibfield{author}{\bibinfo{person}{Peng Wang}, \bibinfo{person}{Shuai Bai}, \bibinfo{person}{Sinan Tan}, \bibinfo{person}{Shijie Wang}, \bibinfo{person}{Zhihao Fan}, \bibinfo{person}{Jinze Bai}, \bibinfo{person}{Keqin Chen}, \bibinfo{person}{Xuejing Liu}, \bibinfo{person}{Jialin Wang}, \bibinfo{person}{Wenbin Ge}, \bibinfo{person}{Yang Fan}, \bibinfo{person}{Kai Dang}, \bibinfo{person}{Mengfei Du}, \bibinfo{person}{Xuancheng Ren}, \bibinfo{person}{Rui Men}, \bibinfo{person}{Dayiheng Liu}, \bibinfo{person}{Chang Zhou}, \bibinfo{person}{Jingren Zhou}, {and} \bibinfo{person}{Junyang Lin}.} \bibinfo{year}{2024}\natexlab{}.
\newblock \bibinfo{title}{Qwen2-{VL}: {Enhancing} {Vision}-{Language} {Model}'s {Perception} of the {World} at {Any} {Resolution}}.
\newblock
\href{https://doi.org/10.48550/arXiv.2409.12191}{doi:\nolinkurl{10.48550/arXiv.2409.12191}}
\newblock
\shownote{arXiv:2409.12191 [cs]}.


\bibitem[Wei et~al\mbox{.}(2024)]%
        {wei_general_2024}
\bibfield{author}{\bibinfo{person}{Haoran Wei}, \bibinfo{person}{Chenglong Liu}, \bibinfo{person}{Jinyue Chen}, \bibinfo{person}{Jia Wang}, \bibinfo{person}{Lingyu Kong}, \bibinfo{person}{Yanming Xu}, \bibinfo{person}{Zheng Ge}, \bibinfo{person}{Liang Zhao}, \bibinfo{person}{Jianjian Sun}, \bibinfo{person}{Yuang Peng}, \bibinfo{person}{Chunrui Han}, {and} \bibinfo{person}{Xiangyu Zhang}.} \bibinfo{year}{2024}\natexlab{}.
\newblock \bibinfo{title}{General {OCR} {Theory}: {Towards} {OCR}-2.0 via a {Unified} {End}-to-end {Model}}.
\newblock
\href{https://doi.org/10.48550/arXiv.2409.01704}{doi:\nolinkurl{10.48550/arXiv.2409.01704}}
\newblock
\shownote{arXiv:2409.01704 [cs]}.


\bibitem[Wu et~al\mbox{.}(2024)]%
        {wu_longvideobench_2024}
\bibfield{author}{\bibinfo{person}{Haoning Wu}, \bibinfo{person}{Dongxu Li}, \bibinfo{person}{Bei Chen}, {and} \bibinfo{person}{Junnan Li}.} \bibinfo{year}{2024}\natexlab{}.
\newblock \showarticletitle{{LongVideoBench}: {A} {Benchmark} for {Long}-context {Interleaved} {Video}-{Language} {Understanding}}.
\newblock \bibinfo{journal}{\emph{Advances in Neural Information Processing Systems}}  \bibinfo{volume}{37} (\bibinfo{date}{Dec.} \bibinfo{year}{2024}), \bibinfo{pages}{28828--28857}.
\newblock
\urldef\tempurl%
\url{https://proceedings.neurips.cc/paper_files/paper/2024/hash/329ad516cf7a6ac306f29882e9c77558-Abstract-Datasets_and_Benchmarks_Track.html}
\showURL{%
\tempurl}


\bibitem[Yuan et~al\mbox{.}(2021)]%
        {yuan_bartscore_2021}
\bibfield{author}{\bibinfo{person}{Weizhe Yuan}, \bibinfo{person}{Graham Neubig}, {and} \bibinfo{person}{Pengfei Liu}.} \bibinfo{year}{2021}\natexlab{}.
\newblock \showarticletitle{{BARTScore}: {Evaluating} {Generated} {Text} as {Text} {Generation}}. In \bibinfo{booktitle}{\emph{Advances in {Neural} {Information} {Processing} {Systems}}}, Vol.~\bibinfo{volume}{34}. \bibinfo{publisher}{Curran Associates, Inc.}, \bibinfo{pages}{27263--27277}.
\newblock
\urldef\tempurl%
\url{https://proceedings.neurips.cc/paper_files/paper/2021/hash/e4d2b6e6fdeca3e60e0f1a62fee3d9dd-Abstract.html}
\showURL{%
\tempurl}


\bibitem[Zaheer et~al\mbox{.}(2020)]%
        {zaheer_big_2020}
\bibfield{author}{\bibinfo{person}{Manzil Zaheer}, \bibinfo{person}{Guru Guruganesh}, \bibinfo{person}{Kumar~Avinava Dubey}, \bibinfo{person}{Joshua Ainslie}, \bibinfo{person}{Chris Alberti}, \bibinfo{person}{Santiago Ontanon}, \bibinfo{person}{Philip Pham}, \bibinfo{person}{Anirudh Ravula}, \bibinfo{person}{Qifan Wang}, \bibinfo{person}{Li Yang}, {and} \bibinfo{person}{Amr Ahmed}.} \bibinfo{year}{2020}\natexlab{}.
\newblock \showarticletitle{Big {Bird}: {Transformers} for {Longer} {Sequences}}. In \bibinfo{booktitle}{\emph{Advances in {Neural} {Information} {Processing} {Systems}}}, Vol.~\bibinfo{volume}{33}. \bibinfo{publisher}{Curran Associates, Inc.}, \bibinfo{pages}{17283--17297}.
\newblock
\urldef\tempurl%
\url{https://papers.neurips.cc/paper_files/paper/2020/hash/c8512d142a2d849725f31a9a7a361ab9-Abstract.html}
\showURL{%
\tempurl}


\bibitem[Zhai et~al\mbox{.}(2023)]%
        {zhai_sigmoid_2023}
\bibfield{author}{\bibinfo{person}{Xiaohua Zhai}, \bibinfo{person}{Basil Mustafa}, \bibinfo{person}{Alexander Kolesnikov}, {and} \bibinfo{person}{Lucas Beyer}.} \bibinfo{year}{2023}\natexlab{}.
\newblock \showarticletitle{Sigmoid {Loss} for {Language} {Image} {Pre}-{Training}}. In \bibinfo{booktitle}{\emph{2023 {IEEE}/{CVF} {International} {Conference} on {Computer} {Vision} ({ICCV})}}. \bibinfo{pages}{11941--11952}.
\newblock
\href{https://doi.org/10.1109/ICCV51070.2023.01100}{doi:\nolinkurl{10.1109/ICCV51070.2023.01100}}
\newblock
\shownote{ISSN: 2380-7504}.


\bibitem[Zhang et~al\mbox{.}(2024b)]%
        {zhang_systematic_2024}
\bibfield{author}{\bibinfo{person}{Haopeng Zhang}, \bibinfo{person}{Philip~S. Yu}, {and} \bibinfo{person}{Jiawei Zhang}.} \bibinfo{year}{2024}\natexlab{b}.
\newblock \bibinfo{title}{A {Systematic} {Survey} of {Text} {Summarization}: {From} {Statistical} {Methods} to {Large} {Language} {Models}}.
\newblock
\href{https://doi.org/10.48550/arXiv.2406.11289}{doi:\nolinkurl{10.48550/arXiv.2406.11289}}
\newblock
\shownote{arXiv:2406.11289 [cs] version: 1}.


\bibitem[Zhang et~al\mbox{.}(2024c)]%
        {zhang_long_2024}
\bibfield{author}{\bibinfo{person}{Peiyuan Zhang}, \bibinfo{person}{Kaichen Zhang}, \bibinfo{person}{Bo Li}, \bibinfo{person}{Guangtao Zeng}, \bibinfo{person}{Jingkang Yang}, \bibinfo{person}{Yuanhan Zhang}, \bibinfo{person}{Ziyue Wang}, \bibinfo{person}{Haoran Tan}, \bibinfo{person}{Chunyuan Li}, {and} \bibinfo{person}{Ziwei Liu}.} \bibinfo{year}{2024}\natexlab{c}.
\newblock \showarticletitle{Long {Context} {Transfer} from {Language} to {Vision}}.
\newblock \bibinfo{journal}{\emph{CoRR}} (\bibinfo{date}{Jan.} \bibinfo{year}{2024}).
\newblock
\urldef\tempurl%
\url{https://openreview.net/forum?id=VZyBATUZjr}
\showURL{%
\tempurl}


\bibitem[Zhang et~al\mbox{.}(2024a)]%
        {zhang_cross-modal_2024}
\bibfield{author}{\bibinfo{person}{Xiang Zhang}, \bibinfo{person}{Senyu Li}, \bibinfo{person}{Ning Shi}, \bibinfo{person}{Bradley Hauer}, \bibinfo{person}{Zijun Wu}, \bibinfo{person}{Grzegorz Kondrak}, \bibinfo{person}{Muhammad Abdul-Mageed}, {and} \bibinfo{person}{Laks V.~S. Lakshmanan}.} \bibinfo{year}{2024}\natexlab{a}.
\newblock \bibinfo{title}{Cross-{Modal} {Consistency} in {Multimodal} {Large} {Language} {Models}}.
\newblock
\href{https://doi.org/10.48550/arXiv.2411.09273}{doi:\nolinkurl{10.48550/arXiv.2411.09273}}
\newblock
\shownote{arXiv:2411.09273 [cs]}.


\bibitem[Zhu et~al\mbox{.}(2024)]%
        {zhu_unraveling_2024}
\bibfield{author}{\bibinfo{person}{Tinghui Zhu}, \bibinfo{person}{Qin Liu}, \bibinfo{person}{Fei Wang}, \bibinfo{person}{Zhengzhong Tu}, {and} \bibinfo{person}{Muhao Chen}.} \bibinfo{year}{2024}\natexlab{}.
\newblock \bibinfo{title}{Unraveling {Cross}-{Modality} {Knowledge} {Conflicts} in {Large} {Vision}-{Language} {Models}}.
\newblock
\href{https://doi.org/10.48550/arXiv.2410.03659}{doi:\nolinkurl{10.48550/arXiv.2410.03659}}
\newblock
\shownote{arXiv:2410.03659 [cs]}.


\bibitem[Zhu and Bhat(2020)]%
        {zhu_gruen_2020}
\bibfield{author}{\bibinfo{person}{Wanzheng Zhu} {and} \bibinfo{person}{Suma Bhat}.} \bibinfo{year}{2020}\natexlab{}.
\newblock \showarticletitle{{GRUEN} for {Evaluating} {Linguistic} {Quality} of {Generated} {Text}}. In \bibinfo{booktitle}{\emph{Findings of the {Association} for {Computational} {Linguistics}: {EMNLP} 2020}}, \bibfield{editor}{\bibinfo{person}{Trevor Cohn}, \bibinfo{person}{Yulan He}, {and} \bibinfo{person}{Yang Liu}} (Eds.). \bibinfo{publisher}{Association for Computational Linguistics}, \bibinfo{address}{Online}, \bibinfo{pages}{94--108}.
\newblock
\href{https://doi.org/10.18653/v1/2020.findings-emnlp.9}{doi:\nolinkurl{10.18653/v1/2020.findings-emnlp.9}}


\end{thebibliography}

\appendix

\end{document}